\documentclass[sigconf]{acmart}
\usepackage{dsfont}

\usepackage{amssymb}
\usepackage{algpseudocode}

\usepackage{soul}
\usepackage{url}
\usepackage[utf8]{inputenc}
\usepackage{caption}
\usepackage{graphicx}
\usepackage{amsmath}
\usepackage{amsthm}
\usepackage{booktabs}
\usepackage{algorithm}

\usepackage{graphicx}
\usepackage{xcolor}
\usepackage{enumitem}
\usepackage{multirow}
\usepackage{subcaption}
\usepackage{tabularx}
\usepackage{balance}

\newtheorem{definition}{Definition}

\newtheorem{assumption}{Assumption}
\newtheorem{problem}{Problem}

\makeatletter
\newcommand{\multiline}[1]{%
  \begin{tabularx}{\dimexpr\linewidth-\ALG@thistlm}[t]{@{}X@{}}
    #1
  \end{tabularx}
}
\makeatother

\AtBeginDocument{%
  }

\setcopyright{acmlicensed}
\copyrightyear{2024}
\acmYear{2024}
\acmDOI{10.1145/3627673.3679727}

\setcopyright{acmlicensed}\acmConference[CIKM '24]{33rd ACM International Conference on
Information and Knowledge Management}{October 21--25, 2024}{Boise, ID, USA}
\acmISBN{979-8-4007-0436-9/24/10}

\acmSubmissionID{1190}

\begin{document}

\title{Learning Fair Invariant Representations under Covariate and Correlation Shifts Simultaneously}
\newcommand{\sysname}{FLAIR}


\author{Dong Li}
\orcid{0000-0003-0081-9318}
\affiliation{%
  \institution{Tianjin University}
  \city{Tianjin}
  \country{China}
}
\email{li13255486422@gmail.com}

\author{Chen Zhao}
\orcid{0000-0002-6400-0048}
\affiliation{%
  \institution{Baylor University}
  \city{Waco, Texas}
  \country{USA}}
\email{chen_zhao@baylor.edu}

\author{Minglai Shao}
\authornotemark[1]
\affiliation{%
  \institution{Tianjin University}
  \city{Tianjin}
  \country{China}}
\email{shaoml@tju.edu.cn}

\author{Wenjun Wang}
\authornote{Corresponding authors}
\affiliation{%
  \institution{Tianjin University}
  \city{Tianjin}
  \country{China}}
\email{13255486722@163.com}

\renewcommand{\shortauthors}{Dong Li, Chen Zhao, Minglai Shao, \& Wenjun Wang}

\begin{abstract}
  Achieving the generalization of an invariant classifier from training domains to shifted test domains while simultaneously considering model fairness is a substantial and complex challenge in machine learning. Existing methods address the problem of fairness-aware domain generalization, focusing on either covariate shift or correlation shift, but rarely consider both at the same time. In this paper, we introduce a novel approach that focuses on learning a fairness-aware domain-invariant predictor within a framework addressing both covariate and correlation shifts simultaneously, ensuring its generalization to unknown test domains inaccessible during training. In our approach, data are first disentangled into content and style factors in latent spaces. Furthermore, fairness-aware domain-invariant content representations can be learned by mitigating sensitive information and retaining as much other information as possible.  Extensive empirical studies on benchmark datasets demonstrate that our approach surpasses state-of-the-art methods with respect to model accuracy as well as both group and individual fairness.
\end{abstract}




\maketitle

\section{Introduction}
\label{sec:introduction}

While machine learning has achieved remarkable success in various areas, including computer vision \cite{krizhevsky2012imagenet}, natural language processing \cite{devlin2018bert}, and many others \cite{jin2020feature, wang2020domainmix, hu2022domain}, these accomplishments are often built upon the assumption that training and test data are independently and identically distributed (\textit{i.i.d.}) within their respective domains \cite{wang2022generalizing}. 
\begin{figure}[t]
    \centering
    \setlength{\abovecaptionskip}{0pt}
    \setlength{\belowcaptionskip}{-10pt}
    \includegraphics[width=1\linewidth]{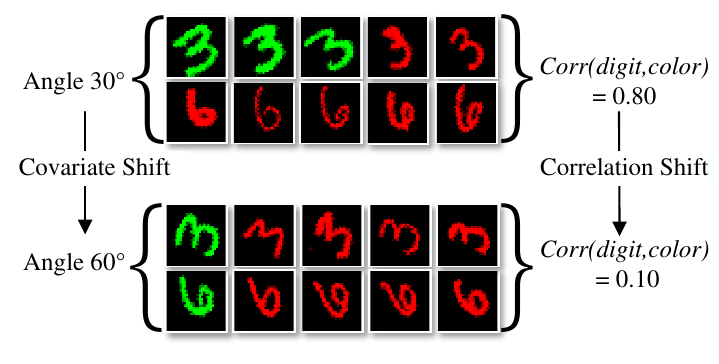}
    \caption{Taking a digit dataset (\textit{e.g.} \texttt{RCMNIST}) as an example to illustrate covariate shift and correlation shift across domains. Here, domain is uniquely determined by the rotation angle and $Corr(digit,color)$, the color serves as the sensitive attribute. $Corr(digit,color)$ represents the correlation between the digit (3 and 6) and color (red and green).}
    \label{fig:intro}

\end{figure}

However, models under this assumption tend to perform poorly when there is a distribution shift between the training and test domains. 
Addressing distribution shifts across domains and generalizing from finite training domains to unseen but related test domains is the primary goal of domain generalization (DG) \cite{IRM}.



Many types of distribution shift are introduced in \cite{survey}, such as label shift \cite{labelshift}, concept shift \cite{conceptshift}, covariate shift \cite{shimodaira2000improving}, and correlation shift \cite{roh2023improving}.
The covariate shift is defined as the differences in the marginal distributions over instances across different domains \cite{shimodaira2000improving}. 
As shown in Figure \ref{fig:intro}, the two domains exhibit variations resulting from different image styles, represented by varying rotation angles.
Correlation shift is defined as the variation in the dependency between the sensitive attribute and label across domains. For example, in Figure \ref{fig:intro}, it is evident that there is a strong correlation between the digit (3,6) and digit colors (green, red) when rotated at $30^{\circ}$, whereas this correlation becomes less pronounced at $60^{\circ}$.

Since the correlation involves sensitive attributes, correlation shift is highly related to fairness.
In the context of algorithmic decision-making, fairness means the absence of any bias or favoritism towards an individual or group based on their inherent or acquired characteristics \cite{mehrabi2021survey}. Many methods have been proposed to address the domain generalization (DG) problem \cite{IRM,MLDG,li2023contrastive,CORAL,DDG}, but most of them lack fairness considerations. Therefore, when these algorithms are applied in human-centered real-world settings, they may exhibit bias against populations \cite{kang2021fair} characterized by sensitive features, such as gender and race. 


While existing efforts have addressed the challenge of fairness-aware domain generalization due to shifted domains, they either overlook the variation in data across domains in the marginal distribution of data features \cite{EIIL,FarconVAE} or specifically address the spurious correlation between sensitive attributes and predicted outcomes in terms of unchanged group fairness \cite{FATDM} across domains. Therefore, research is needed to explore fairness-aware domain generalization considering both covariate and correlation shifts simultaneously across training and test domains.

\begin{figure}[t]
    \centering
    \setlength{\abovecaptionskip}{0pt}
    \setlength{\belowcaptionskip}{-10pt}
    \includegraphics[width=\linewidth]{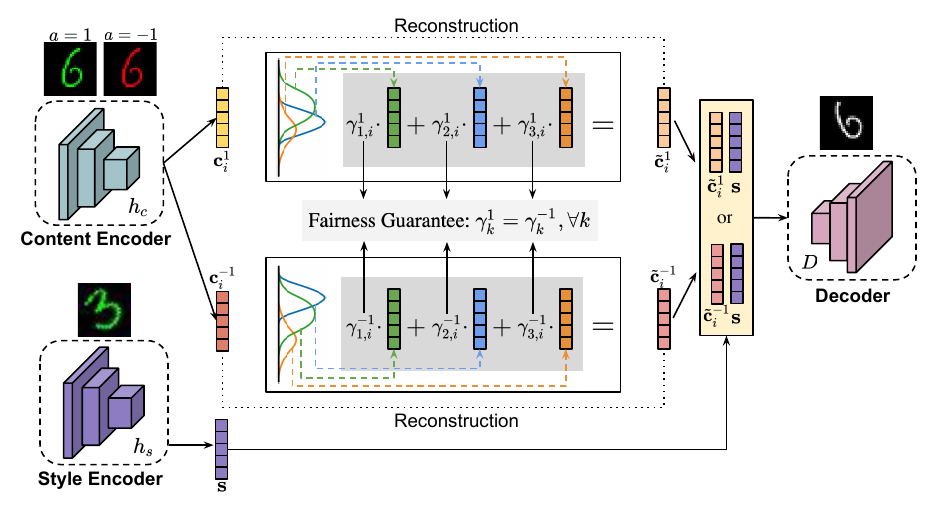}
    \caption{Illustrating the pipeline of \sysname{} using \texttt{RCMNIST} dataset as an example. The content encoder $h_c$ first maps instances to the latent content space to obtain latent content factors. Subsequently, these content factors are grouped based on the sensitive attributes (color) into $\textbf{c}_i^1$ and $\textbf{c}_i^{-1}$. Consequently, the fair content representations $\tilde{\textbf{c}}_i^1$ and $\tilde{\textbf{c}}_i^{-1}$ are reconstructed using weighted prototypes. Each prototype represents a statistical mean estimated from its corresponding cluster, which is fitted by the content factors of the respective subgroups, while ensuring fairness through Eq.(\ref{eq:fair_constraint}). Further, instances are transformed into different domains using the style factor $\textbf{s}$ extracted by the style encoder $h_s$.}
    \label{fig:model}
\end{figure}

In this paper, we propose a novel framework, namely Fairness-aware LeArning Invariant Representations (\sysname{}). It focuses on the problem arising from both covariate shift and correlation shift while considering fairness. The overall framework is shown in Figure \ref{fig:model}.
In the presence of multiple training domains, our objective is to acquire a predictor that is both domain-invariant and fairness-aware. This enables effective generalization in unseen test domains while preserving both accuracy and fairness.
We assume there exists an underlying transformation model that can transform instances sampled from one domain to another while keeping the class labels unchanged. 
Under this assumption, the predictor consists of three components: a content featurizer, a fair representation learner, and an invariant classifier. 
To achieve fairness, data are divided into different sensitive subgroups. Within each subgroup, content factors encoded from the content featurizer are reconstructed using $K$ latent prototypes.
These reconstructed content representations over various sensitive subgroups are crafted with dual objectives: (1) minimizing the inclusion of sensitive information and (2) maximizing the preservation of non-sensitive information.
Utilizing these representations as inputs, we train a fairness-aware domain-invariant classifier for making model predictions.
Exhausted experiments showcase that \sysname{} demonstrates robustness in the face of covariate shift, even when facing alterations in unfairness and correlation shift across domains.
The main contributions are summarized:
\begin{itemize}[leftmargin=*]
    \item We introduce a fairness-aware domain generalization problem within a framework  that addresses both covariate and correlation shifts simultaneously, which has practical significance.
    \item We introduce an end-to-end training approach aimed at learning a fairness-aware domain invariant predictor. 
    We claim that the trained predictor can generalize effectively to unseen test domains that are unknown and inaccessible during training.
    \item Comprehensive experiments on three benchmark datasets
    show that our proposed algorithm \sysname{} significantly outperforms state-of-the-art baselines with respect to model accuracy as well as both group and individual fairness.
\end{itemize}

\section{Related Work}
\label{sec:relatedwork}
\textbf{Algorithmic Fairness in Machine Learning.}
In recent years, fairness in machine learning has gained widespread attention. In this field, there is a widely recognized trade-off: enhancing fairness may come at the cost of accuracy to some extent \cite{chen2018my,menon2018cost}. How to handle such a trade-off, especially in real-world datasets, has been a widely researched issue in the field of algorithmic fairness.



From a statistical perspective, algorithmic fairness metrics are typically divided into group fairness and individual fairness. The conflict between them is a common challenge, as algorithms that achieve group fairness may not be able to handle individual fairness \cite{ifair}. LFR \cite{LFR} is the first method to achieve both group fairness and individual fairness simultaneously. It encodes tabular data, aiming to preserve the original data as much as possible while ignoring information related to sensitive attributes.

\textbf{Fairness-Aware Domain Generalization.}
Some efforts \cite{zhao2023towards,zhao2022adaptive,zhao2021fairnessphd,zhao2024dynamic,zhao2021fairness} have already been attempted to address the fairness-aware domain generalization problem. EIIL \cite{EIIL} takes correlation shift into consideration when addressing the DG problem, thus ensuring fairness to some extent. FVAE \cite{FarconVAE} learns fair representation through contrastive learning and both improve out-of-distribution generalization and fairness. But both of them only take correlation shift into account while assuming that covariate shift remains invariant. The latest work FATDM \cite{FATDM} attempts to simultaneously enhance the model's accuracy and fairness, considering the DG problem associated with covariate shift. However, it does not consider correlation shift and solely focuses on group fairness, without addressing individual fairness.


\section{Preliminaries}
\label{sec:preliminaries}
\textbf{Notations.}
Let $\mathcal{X}\subseteq\mathbb{R}^d$ denote a feature space, $\mathcal{A}=\{-1,1\}$ is a sensitive space, and $\mathcal{Y}=\{0,1\}$ is a label space for classification. Let $\mathcal{C}\subseteq\mathbb{R}^c$ and $\mathcal{S}\subseteq\mathbb{R}^s$ be the latent content and style spaces, respectively, induced from $\mathcal{X}$ by an underlying transformation model $T:\mathcal{X}\times\mathcal{X}\rightarrow\mathcal{X}$. We use $X,A,Y,C,S$ to denote random variables that take values in $\mathcal{X,A,Y,C,S}$ and $\mathbf{x},a,y,\mathbf{c},\mathbf{s}$ be the realizations. A domain $e\in\mathcal{E}$ is specified by distribution $\mathbb{P}(X^e,A^e,Y^e):\mathcal{X}\times\mathcal{A}\times\mathcal{Y}\rightarrow[0,1]$. A predictor $f$ parameterized by $\boldsymbol{\theta}_{cls}$ denotes $f:\mathcal{X}\times\mathcal{A}\times\Theta\rightarrow\mathcal{Y}$.

\textbf{Problem Formulation.}
We consider a set of data domains $\mathcal{E}$, where each domain $e\in\mathcal{E}$ corresponds to a distinct data $\mathcal{D}^e=\{(\mathbf{x}^e_i,a^e_i,y^e_i)\}_{i=1}^{|\mathcal{D}^e|}$ sampled \textit{i.i.d.} from $\mathbb{P}(X^e,A^e,Y^e)$. 
Given a dataset $\mathcal{D}=\{\mathcal{D}^e\}_{e\in\mathcal{E}}$, it is partitioned into 
a training dataset $\mathcal{D}_{tr}\subset\mathcal{D}$ with multiple training domains $\mathcal{E}_{tr}\subset\mathcal{E}$ and a test dataset $\mathcal{D}_{te}=\mathcal{D}\backslash\mathcal{D}_{tr}$ with unknown test domains which are inaccessible during training. Therefore, given samples from finite training domains, we aim to learn a fairness-aware predictor $f$ at training that is generalizable on unseen test domains.
\begin{problem}[Domain generalization concerning fairness]
\label{prob:problem}
    Let $\mathcal{E}_{tr}\subset\mathcal{E}$ be a finite subset of training domains and assume that for each $e\in\mathcal{E}_{tr}$, we have access to its corresponding data $\mathcal{D}^e=\{(\mathbf{x}^e_i,a^e_i,y^e_i)\}_{i=1}^{|\mathcal{D}^e|}$ sampled \textit{i.i.d.} from $\mathbb{P}(X^e,A^e,Y^e)$. Given a loss function $\ell_{CE}:\mathcal{Y}\times\mathcal{Y}\rightarrow\mathbb{R}$, the goal is to learn a fair predictor $f$ parameterized by $\boldsymbol{\theta}_{cls}\in\Theta_{fair}\subset \Theta$ for any $\mathcal{D}^e\in\mathcal{D}_{tr}$ that minimizes the worst-case risk over training domains $\mathcal{E}_{tr}$ that 
    \begin{equation*}
        \begin{aligned}
        \min_{\boldsymbol{\theta}_{cls}\in\Theta_{fair}}\max_{e\in\mathcal{E}_{tr}} \:\: \mathbb{E}_{\mathbb{P}(X^e,A^e,Y^e)}\ell_{CE}(f(X^e,A^e,\boldsymbol{\theta}_{cls}),Y^e)
    \end{aligned}
    \end{equation*}
\end{problem}
However, addressing Problem \ref{prob:problem} by training such a predictor $f$ is challenging because (1) $f$ is required to remain invariant across domains in terms of model accuracy, and model outcomes are fair with respect to sensitive subgroups defined by $A$; and (2) we do not assume data from $\mathcal{E}\backslash\mathcal{E}_{tr}$ is accessible during training.

To tackle such challenges, we divide the fairness-aware domain invariant predictor $f$ into three components: a domain-invariant featurizer $h_c:\mathcal{X}\times\Theta_{fair}\rightarrow\mathcal{C}$ parameterized by $\boldsymbol{\theta}_c$, a fair representation learner $g:\mathcal{C}\times\mathcal{A}\times\Theta_{fair}\rightarrow\mathcal{C}$ parameterized by $\boldsymbol{\theta}_g$, and a classifier $\omega:\mathcal{C}\times\Theta_{fair}\rightarrow\mathbb{R}$ parameterized by $\boldsymbol{\theta}_w$, denoted $f=h_c\circ g\circ \omega$ and $\boldsymbol{\theta}_{cls}=\{\boldsymbol{\theta}_c,\boldsymbol{\theta}_g,\boldsymbol{\theta}_w\}$.

\section{Fairness-aware Learning Invariant Representations (\sysname{})}
\label{sec:method}
In this paper, we narrow the scope of various distribution shifts and focus on a hybrid shift where covariate and correlation shifts are present simultaneously.

\begin{definition}[Covariate shift and correlation shift]
\label{def:shifts}
    Given $\forall e_1, e_2\in\mathcal{E}$ and $e_1\ne e_2$, a covariate shift occurs in Problem \ref{prob:problem} when domain variation is due to differences in the marginal distributions over input features $\mathbb{P}(X^{e_1}) \ne \mathbb{P}(X^{e_2})$. 
    Meanwhile, a correlation shift arises in Problem \ref{prob:problem} when domain variation results from changes in the joint distribution between $Y$ and $Z$, denoted as $\mathbb{P}(A^{e_1},Y^{e_1}) \ne \mathbb{P}(A^{e_2},Y^{e_2})$. More specifically, $\mathbb{P}(Y^{e_1}|A^{e_1}) \ne \mathbb{P}(Y^{e_2}|A^{e_2})$ and $\mathbb{P}(A^{e_1}) = \mathbb{P}(A^{e_2})$; or $\mathbb{P}(A^{e_1}|Y^{e_1}) \ne \mathbb{P}(A^{e_2}|Y^{e_2})$ and $\mathbb{P}(Y^{e_1}) = \mathbb{P}(Y^{e_2})$.
\end{definition}

In Section \ref{sec:ddv}, we handle covariate shift by enforcing invariance on instances based on disentanglement, while in Section \ref{sec:LFCR}, we address correlation shift by learning fair content representation.

\subsection{Disentanglement of Domain Variation}
\label{sec:ddv}
In \cite{MBDG}, distribution shifts are attributed into two forms: concept shift, where the distribution of instance classes varies across different domains, and covariate shift, where the marginal distributions over instance $\mathbb{P}(X^e)$ are various. In this paper, we restrict the scope of our framework to focus on Problem \ref{prob:problem} in which inter-domain variation is solely due to covariate shift.

Building upon the insights from existing domain generalization literature \cite{DDG,MBDG,zhao2023towards}, data variations across domains are disentangled into multiple factors in latent spaces.
\begin{assumption}[Latent Factors]
\label{assump:latentfactors}
    Given $\mathcal{D}^e=\{(\mathbf{x}_i^e,a_i^e,y_i^e)\}_{i=1}^{|\mathcal{D}^e|}$ sampled \textit{i.i.d.} from $\mathbb{P}(X^e,A^e,Y^e)$ in domain $e\in\mathcal{E}$, we assume that each instance $(\mathbf{x}_i^e,a_i^e,y_i^e)$ is generated from
    \begin{itemize}[leftmargin=*]
        \item a latent content factor $\mathbf{c}=h_c(\mathbf{x}_i^e,\boldsymbol{\theta}_c)\in\mathcal{C}$, where $\mathcal{C}=\{\mathbf{c}_{y=0}, \mathbf{c}_{y=1}\}$ refers to a content space, and $h_c$ is a content encoder;
        \item a latent style factor $\mathbf{s}^e=h_s(\mathbf{x}_i^e, \boldsymbol{\theta}_s)\in\mathcal{S}$, where $\mathbf{s}^e$ is specific to the individual domain $e$, and $h_s:\mathcal{X}\times\Theta\rightarrow\mathcal{S}$ is a style encoder.
    \end{itemize}
    We assume that the content factors in $\mathcal{C}$ do not change across domains. Each domain $e$ over $\mathbb{P}(X^e,A^e,Y^e)$ is represented by a unique $\mathbf{s}^e$ and $Corr(Y^e,A^e)$, where $Corr(Y^e,A^e)$ is the correlation betweem $Y^e$ and $A^e$.
\end{assumption}

Under Assumption \ref{assump:latentfactors}, we further assume that, for any two domains $e_i,e_j\in\mathcal{E}$, inter-domain variations between them due to covariate shift are managed via an underlying transformation model $T$. Through this model, instances sampled from such two domains can be transformed interchangeably.

\begin{assumption}[Transformation Model]
\label{assump:T}
    We assume, $\forall e,e'\in\mathcal{E}, e\neq e'$, there exists a function $T:\mathcal{X}\times\mathcal{X}\rightarrow\mathcal{X}$ that transforms instances from domain $e$ to $e'$, denoted as $X^{e'}=T(X^{e},X^{e'})$. The transformation model $T$ is defined as
    \begin{equation*}
        \begin{aligned}
        T(X^{e},X^{e'})=D(h_c(X^{e},\boldsymbol{\theta}_c),h_s(X^{e'},\boldsymbol{\theta}_s),\boldsymbol{\theta}_d )    
    \end{aligned}
    \end{equation*}
    where $h_c$ and $h_s$ are content and style encoders defined in Assumption \ref{assump:latentfactors}, and $D:\mathcal{C}\times\mathcal{S}\times\Theta\rightarrow\mathcal{X}$ denotes a decoder.
\end{assumption}

With the transformation model $T$ that transforms instances from domain $e$ to $e'$, $\forall e,e'\in\mathcal{E}$, under Assumption \ref{assump:T}, we introduce a new definition of invariance with respect to the variation captured by $T$ in Definition \ref{def:Tinvariance}.

\begin{definition}[$T$-invariance]
\label{def:Tinvariance}
    Under Assumptions \ref{assump:latentfactors} and \ref{assump:T}, given a transformation model $T$ as well as two instance $(\mathbf{x}^{e}_i,a^{e}_i,y^{e}_i)$ and $(\mathbf{x}^{e'}_j,a^{e'}_j,y^{e'}_j)$, a content encoder $h_c$ is domain invariant if it holds
    \begin{equation}
    \begin{aligned}
    \label{eq:Lossinv}
        \mathbf{x}^{e'}_j=T(\mathbf{x}^{e}_i,\mathbf{x}^{e'}_j)&,\quad \text{when} \quad e\neq e', y^{e}=y^{e'}, \quad \text{or}\\
        \mathbf{x}^{e}_i=T(\mathbf{x}^{e}_i,\mathbf{x}^{e'}_j)&,\quad \text{when} \quad e = e', y^{e}\neq y^{e'}
    \end{aligned}
    \end{equation}
    almost surely $\forall e,e'\in\mathcal{E}$. 
\end{definition}
Definition \ref{def:Tinvariance} is crafted to enforce invariance on instances based on disentanglement via $T$. 
The output of $h_c$ is further utilized to acquire a fairness-aware representation, considering different sensitive subgroups, through the learner $g$ within the content latent space.

\subsection{Learning Fair Content Representations}
\label{sec:LFCR}

\textit{Dwork et al.,} \cite{dwork2012fairness} defines fairness that \textit{similar individuals are treated similarly}.
As stated in Section \ref{sec:ddv}, the featurizer $h_c$ maps instances to the latent content space. 
Therefore, for each instance $(\mathbf{x}_i^e,a_i^e,y_i^e)$ sampled \textit{i.i.d.} from $\mathbb{P}(X^e,A^e,Y^e)$ where $e\in\mathcal{E}_{tr}$, the goal of the learner $g$ is to reconstruct a fair content representation $\Tilde{\mathbf{c}}_i=g(\mathbf{c}_i,\boldsymbol{\theta}_g)$ from $\mathbf{c}_i=h_c(\mathbf{x}_i^e,\boldsymbol{\theta}_c)$, wherein $\Tilde{\mathbf{c}}_i$ is generated to meet two objectives (1) minimizing the information disclosure related to a specific sensitive subgroup $\mathcal{D}_{tr}^{a=-1}$ or $\mathcal{D}_{tr}^{a=1}$, and (2) maximizing the preservation of significant information within non-sensitive representations. 
Under Assumption \ref{assump:latentfactors}, since the content space is invariant across domains, we omit the superscript of domain labels for content factors. 

To achieve these objectives effectively through $g$ and drawing inspiration from \cite{LFR,ifair}, we group the content factors along with the sensitive attributes, denoted $\{\mathbf{c}_i^a\}_{i=1}^{N^a}= \{(\mathbf{c}_i,a_i)\}_{i=1}^{N^a}$, of instances $\{(\mathbf{x}^e_i,a^e_i,y^e_i)\}_{i=1}^{N^a}$ within each sensitive subgroup $\mathcal{D}_{tr}^a,\forall a \in\{-1,1\}$, which are encoded from $h_c$, into $K$ clusters based on their similarity.
Consequently, their fair content representations $\{\Tilde{\mathbf{c}}_i^a\}_{i=1}^{N^a}$, with the sensitive attributes $\{a_i\}_{i=1}^{N^a}$ unchanged, are reconstructed using weighted prototypes, with each prototype $\boldsymbol{\mu}_k^a$ representing a statistical mean estimated from each cluster.


Specifically, for content factors $\{\mathbf{c}_i^a\}_{i=1}^{N^a}$ in a sensitive subgroup $a$ where $a\in\{-1,1\}$, let $Z$ be a latent variable, where its realization $\mathbf{z}^a\in\{0,1\}^K$ is a $K$-dimensional vector, satisfying a particular entry $z_k^a$ is equal to $1$, while all other entries are set to $0$s, and $\sum_k z_{k}^a=1$. 
We denote $\pi_{k}^a$ as the mixing coefficients representing the prior probability of $z_k^a=1$ that $\mathbf{c}_i^a$ belongs to the $k$-th prototype. 
\begin{equation*}
    \begin{aligned}
    \mathbb{P}(z_{k}^a=1)=\pi_{k}^a,\quad 0\leq\pi_{k}^a\leq 1,\quad \sum_{k=1}^K \pi_{k}^a=1
    \end{aligned}
\end{equation*}
In the context of Gaussian mixture models, we assume the conditional distribution $(C^a|Z^a=z_{k}^a)\sim\mathcal{N}(\boldsymbol{\mu}_k^a,\Sigma_k^a)$. To estimate the parameters $\boldsymbol{\theta}_g^a=\{\boldsymbol{\mu}^a_k,\Sigma^a_k,\pi^a_k\}_{k=1}^K$ of the subgroup $a$, we take the loss
\begin{equation}
\begin{aligned}
\label{eq:loss_gmm_a}
    \mathcal{L}_{gmm}(\mathbf{x}_i^a,\boldsymbol{\theta}_c,\boldsymbol{\theta}_g^a) = -\sum_{i=1}^{N^a}\ln\Big\{\sum_{k=1}^K\pi_k^a\mathcal{N}(\mathbf{c}_i^a|
    \boldsymbol{\mu}_k^a,\Sigma_k^a)\Big\} + \sum_{k=1}^K\pi_k^a
\end{aligned}
\end{equation}
Intuitively, the latent variable $Z$ is the key to finding the maximal log-likelihood. We attempt to compute the posterior distribution $\gamma_{k,i}^a$ of $Z$ given the observations $\mathbf{c}_i^a$:
\begin{equation}
    \begin{aligned}
    \label{eq:posterior_dis}
        \gamma_{k,i}^a:=\mathbb{P}(z_k^a=1|\mathbf{c}_i^a) = \frac{\pi_k^a\mathcal{N}(\mathbf{c}_i^a|\boldsymbol{\mu}_k^a,\Sigma_k^a)}{\sum_{j=1}^K \pi_j^a\mathcal{N}(\mathbf{c}_i^a|\boldsymbol{\mu}_j^a,\Sigma_j^a)}
    \end{aligned}
\end{equation}

To achieve fairness, the fundamental idea designing $g$ is to make sure that the probability that a random content factor $\mathbf{c}_i^{a=-1}$ from the sensitive subgroup $a=-1$ mapping to the $k$-th particular prototype $\boldsymbol{\mu}_k^{a=-1}$ is equal to the probability of a random content factor $\mathbf{c}_i^{a=1}$ mapping to the prototype $\boldsymbol{\mu}_k^{a=1}$ from the other sensitive subgroup $a=1$.
\begin{equation}
    \begin{aligned}
    \label{eq:fair_constraint}
        \gamma_k^{a=1} = \gamma_k^{a=-1}, \forall k \quad \text{where } \gamma_k^a = \mathbb{E}_{(\mathbf{x}_i,a,y_i)\sim\mathbb{P}(X,A=a,Y)} \gamma_{k,i}^a
    \end{aligned}
\end{equation}
We hence formulate the loss regarding fairness that 
\begin{equation}
    \begin{aligned}
    \label{eq:loss_fair}
        \mathcal{L}_{fair}(\mathcal{D}_{tr},\boldsymbol{\theta}_c,\boldsymbol{\theta}_g) = \sum_{k=1}^K \Big|\gamma_k^{a=1} - \gamma_k^{a=-1}\Big|
    \end{aligned}
\end{equation}
where $\boldsymbol{\theta}_g=\{\boldsymbol{\theta}_g^{a=-1},\boldsymbol{\theta}_g^{a=1}\}$. Eq.(\ref{eq:loss_fair}) draws inspiration from the group fairness metric, known as the Difference of Demographic Parity (DDP) \cite{lohaus2020too}, which enforces the statistical parity between two sensitive subgroups.

To maximize the non-sensitive information in the reconstructed content representations, the reconstruction loss is defined
\begin{equation}
    \begin{aligned}
    \label{eq:loss_rec}
        \mathcal{L}_{rec}(\mathbf{x}_i^a,\boldsymbol{\theta}_c,\boldsymbol{\theta}_g^a) &= \sum_{i=1}^{|\mathcal{E}_{tr}|} dist[\mathbf{c}_i^a, \Tilde{\mathbf{c}}_i^a], \quad \forall a\in\{-1,1\} \\
        \text{where}\quad \Tilde{\mathbf{c}}^a_i &= g(\mathbf{c}_i^a,\boldsymbol{\theta}_g^a)=\sum_{k=1}^K \gamma_{k,i}^a\cdot\boldsymbol{\mu}_k^a
    \end{aligned}
\end{equation}
where $|\mathcal{E}_{tr}|=N^{a=-1}+N^{a=1}$ and $dist[\cdot,\cdot]:\mathcal{C}\times\mathcal{C}\rightarrow\mathbb{R}$ is the Euclidean distance metric. 

\subsection{Learning the Predictor $\boldsymbol{f}$}
To tackle Problem \ref{prob:problem}, which aims to learn a fairness-aware domain invariant predictor $f$, a crucial element of $f$ is the acquisition of content factors through $h_c$, while simultaneously reducing the sensitive information associated with them through $g$. 
In this subsection, we introduce a framework designed to train $f$ with a focus on both domain invariance and model fairness.

Given training domains $\mathcal{E}_{tr}$, a data batch $\mathcal{Q}=\{(\mathbf{r}_{1},\mathbf{r}_{2},\mathbf{r}_{3},\mathbf{r}_{4})_q\}_{q=1}^Q$ containing multiple quartet instance pairs are sampled from $\mathbb{P}(X^e,A^e,$\\$Y^e)$ and $\mathbb{P}(X^{e'},A^{e'},Y^{e'})$, $\forall e,e'\in\mathcal{E}_{tr}$, where $Q$ denotes the number of quartet pairs in $|\mathcal{Q}|$. Specifically,
\begin{equation*}
    \begin{aligned}
        \mathbf{r}_{1}&=(\mathbf{x}^e_{1},a=-1,y), \quad \text{with class } y \text{ and domain } e \\
        \mathbf{r}_{2}&=(\mathbf{x}^e_{2},a=1,y'), \quad \text{with class } y' \text{ and domain } e \\
        \mathbf{r}_{3}&=(\mathbf{x}^{e'}_{3},a=-1,y), \quad \text{with class } y \text{ and domain } e' \\
        \mathbf{r}_{4}&=(\mathbf{x}^{e'}_{4},a=1,y'), \quad \text{with class } y' \text{ and domain } e'
    \end{aligned}
\end{equation*}
We set $\mathbf{r}_1$ and $\mathbf{r}_2$ (same to $\mathbf{r}_3$ and $\mathbf{r}_4$) share the same domain $e$ but different class label $y$ and $y'$, while $\mathbf{r}_1$ and $\mathbf{r}_3$ (same to $\mathbf{r}_2$ and $\mathbf{r}_4$) share the same class label $y$ but different domains $e$ and $e'$. 
Therefore, $\mathbf{r}_1$ and $\mathbf{r}_2$ are alternative instances with respect to $\mathbf{r}_3$ and $\mathbf{r}_4$ in a different domain, respectively.

Therefore, under Definition \ref{def:Tinvariance} and Eq.(\ref{eq:Lossinv}), we have the invariance loss $R_{inv}$ with respect to $\boldsymbol{\theta}_{inv}=\{\boldsymbol{\theta}_c,\boldsymbol{\theta}_s,\boldsymbol{\theta}_d\}$,
\begin{equation}
\begin{aligned}
\label{eq:inv_loss}
    R_{inv} (\boldsymbol{\theta}_{inv})= 
    \mathbb{E}_{(R1,R2,R3,R4)\in\mathcal{Q}} \Big(d[R_1,T(R_1,R_2)]
    +d[R_3,T(R_3,R_4)]\Big)
\end{aligned}
\end{equation}


Note that in each distance metric $d[\cdot]$ of $R_{inv}$, it compares a pair of instances with the same domain but different classes.

Furthermore, given Eq.(\ref{eq:loss_gmm_a}), Eq.(\ref{eq:loss_rec}) and under Definition \ref{def:Tinvariance}, we have the invariant classification loss with respect to $\boldsymbol{\theta}_{cls}=\{\boldsymbol{\theta}_c,\boldsymbol{\theta}_g,\boldsymbol{\theta}_w\}$,
\begin{equation}
    \begin{aligned}
    \label{eq:r_cls}
        R_{cls}(\boldsymbol{\theta}_{cls}) = R_{cls}(\boldsymbol{\theta}_{cls}^{a=-1}) + R_{cls}(\boldsymbol{\theta}_{cls}^{a=1})
    \end{aligned}
\end{equation}
with 
\begin{equation*}
    \begin{aligned}
        R_{cls}(\boldsymbol{\theta}_{cls}^a)=
        \mathbb{E}_{(R_i,R_j)\in\mathcal{Q}} 
        \Big\{
        d[R_i,T(R_i,R_j)]
        + \mathcal{L}_{gmm}(R_i,\boldsymbol{\theta}_c,\boldsymbol{\theta}_g^a) \\\nonumber+\mathcal{L}_{rec}(R_i,\boldsymbol{\theta}_c,\boldsymbol{\theta}_g^a)
        +\ell_{CE}\Big(\omega(g(h_c(R_i,\boldsymbol{\theta}_c),\boldsymbol{\theta}_g^a),\boldsymbol{\theta}_w),Y\Big) \Big\}
    \end{aligned}
\end{equation*}
where $d:\mathcal{X}\times\mathcal{X}\rightarrow\mathbb{R}$ indicates a distance metric, such as $\ell_1$-norm.
$R_{cls}(\boldsymbol{\theta}_{cls}^{a=-1})$ indicates the empirical risk of instance pairs with the sensitive attribute $a=-1$. Similarly, $R_{cls}(\boldsymbol{\theta}_{cls}^{a=1})$ is the empirical risk of instance pairs with the sensitive attribute $a=1$. Notice that the instance pair $(R_i,R_j)$ in $R_{cls}$ sampled from $\mathcal{Q}$ have the same class label but different domains, such as $(\mathbf{r}_1,\mathbf{r}_3)$ and $(\mathbf{r}_2,\mathbf{r}_4)$.

Finally, the fair loss $R_{fair}$ is defined over the data batch with all sensitive attributes using Eq.(\ref{eq:loss_fair}),
\begin{equation}
    \begin{aligned}
    \label{eq:loss_fair_Q}
        R_{fair}(\boldsymbol{\theta}_c,\boldsymbol{\theta}_g)&=\mathcal{L}_{fair}(Q,\boldsymbol{\theta}_c,\boldsymbol{\theta}_g)
    \end{aligned}
\end{equation}
Therefore the total loss is given
\begin{equation}
    \begin{aligned}
    \label{eq:loss_total}
        R_{total} = R_{cls} + \lambda_1\cdot R_{inv} + \lambda_2\cdot R_{fair}
    \end{aligned}
\end{equation}
where $\lambda_1,\lambda_2>0$ are Lagrangian multipliers.
\subsection{An Effective Algorithm}
\label{sec:algor}

\begin{algorithm}[!t]
\caption{Fairness-aware Learning Invariant Representations (\sysname{})}
\label{alg:algo}

\begin{algorithmic}[1]
\Statex \textbf{Input}: Training dataset $\mathcal{D}_{tr}$, stepsize $\eta_1,\eta_2, \eta_3$, margin $\epsilon_1, \epsilon_2$, number of prototypes $K$
\Statex \textbf{Initialize}: primal variables $\boldsymbol{\theta}=\{\boldsymbol{\theta}_c, \boldsymbol{\theta}_s, \boldsymbol{\theta}_d, \boldsymbol{\theta}_g, \boldsymbol{\theta}_w\}$ and dual variables $\lambda_1$, $\lambda_2$
\Repeat
    
    \State Sample a batch $\mathcal{Q}=\{(\mathbf{r}_{1},\mathbf{r}_{2},\mathbf{r}_{3},\mathbf{r}_{4})_q\}_{q=1}^Q$ in $\mathcal{D}_{tr}$.
    \State Evaluate $R_{inv}(\boldsymbol{\theta}_c,\boldsymbol{\theta}_s,\boldsymbol{\theta}_d)$ using $\mathcal{Q}$ and Eq.(\ref{eq:inv_loss}).
    \State Estimate $\boldsymbol{\theta}_g=\Call{FairGMMs}{\mathcal{Q},\boldsymbol{\theta}_c}$
    \State Estimate $\hat{R}_{fair}(\boldsymbol{\theta}_c,\boldsymbol{\theta}_g)\approx\sum_{k=1}^K{\left|\pi_k^{a=-1}-\pi_k^{a=1}\right|}$
    \State Evaluate $R_{cls}(\boldsymbol{\theta}_c,\boldsymbol{\theta}_g,\boldsymbol{\theta}_w)$ using Eq.(\ref{eq:r_cls})
    \State Define $R_{total}$ using Eq.(\ref{eq:loss_total})
    
    \State Primal Update $\boldsymbol{\theta} \gets \text{Adam}(R_{cls}+\lambda_1R_{inv}+\lambda_2\hat{R}_{fair}, \boldsymbol{\theta}, \eta_1)$
    \State Dual Update $\lambda_1\leftarrow\max\Big\{\Big[\lambda_1+\eta_2\Big(R_{inv}-\epsilon_1 \Big)\Big],0\Big\}$,
    \Statex \quad \quad \quad \quad \quad \quad \quad$\lambda_2\leftarrow\max\Big\{\Big[\lambda_2+\eta_3\Big(\hat{R}_{fair}-\epsilon_2 \Big)\Big],0\Big\}$ 
\Until{convergence}
\Procedure{FairGMMs}{$\mathcal{Q},\boldsymbol{\theta}_c$}
        \For{each $a\in\{-1,1\}$}
            \State \multiline{%
            Define $\{\mathbf{c}_i^a\}_{i=1}^{N^a}$ by encoding $\mathcal{Q}$ using $h_c$ and $\boldsymbol{\theta}_c$ with respect to the sensitive subgroup $a$
            }
            \Repeat
                \State Estimate $\{\gamma_{k,i}^a\}_{i=1}^{N^a}$ using Eq.(\ref{eq:posterior_dis})
                \State Update $\boldsymbol{\mu}_k^a\gets\frac{\sum_{i=1}^{N^a}\gamma_{k,i}^a\textbf{c}_i^a}{\sum_{i=1}^{N^a}\gamma_{k,i}^a} $
                \State Update $\Sigma_k^a \gets \frac{\sum_{i=1}^{N^a}\gamma_{k,i}^a(\textbf{c}_i^a-\boldsymbol{\mu}_k^a)(\textbf{c}_i^a-\boldsymbol{\mu}_k^a)^T}{\sum_{i=1}^{N^a}\gamma_{k,i}^a}$
                \State Update $\pi_k^a\gets\left \{
                                \begin{aligned}
                                \frac{\sum_{i=1}^{N^a}\gamma_{k,i}^a}{N^a+\lambda_2},\ \   & \text{if } \pi_k^a \ge \pi_k^{-a}\\
                                \frac{\sum_{i=1}^{N^a}\gamma_{k,i}^a}{N^a-\lambda_2},\ \  & \text{otherwise}
                                \end{aligned}
                                \right.$
            \Until{convergence}
        \EndFor
    \EndProcedure
\end{algorithmic}
\end{algorithm}


We introduce an effective algorithm for \sysname{} to implement the predictor $f$, as shown in Algorithm \ref{alg:algo}. Lines 2-3 represent the transformation model $T$, while lines 4-6 denote the fair representation learner $g$. 
In the $g$ component, we employ $\hat{R}_{fair}$ as an approximation to $R_{fair}$, since the EM algorithm\cite{mclachlan2007algorithm} in \textsc{FairGMMs} continuously estimates $\gamma_k^a$ using $\pi_k^a$, $\forall k,a$.
Parameters of $\boldsymbol{\theta}_g$ update are given in lines 15-18 of Algorithm \ref{alg:algo}. 
We optimize $\lambda_1$ and $\lambda_2$ in the $R_{total}$ using the primal-dual algorithm, which is an effective tool for enforcing invariance \cite{MBDG}. The time complexity of Algorithm \ref{alg:algo} is $\mathcal{O}(M \times Q \times (N^{a=1}+N^{a=-1}))$, where $M$ is the number of batches.

\section{Experimental Settings}
\label{sec:exp}

\begin{table*}[t]
\scriptsize
    \centering
    \setlength{\abovecaptionskip}{0pt}
    \setlength{\belowcaptionskip}{2pt}
    \setlength\tabcolsep{3pt}
    \caption{Performance on \texttt{RCMNIST} (bold is the best, underline is the second best).}
    \label{tab:result-rcmnist}
    \begin{tabular}{l|c|c|c}
        \toprule
        \multirow{2}{*}{} & \multicolumn{3}{c}{Consisitency $\uparrow$ / $\Delta_{DP}$ $\downarrow$ / AUC$_{fair}$ $\downarrow$ / Accuracy $\uparrow$} \\ 
         \cmidrule(lr){2-4}
         & $0^{\circ}$ & $15^{\circ}$ & $30^{\circ}$\\
        \cmidrule(lr){1-1} \cmidrule(lr){2-2} \cmidrule(lr){3-3} \cmidrule(lr){4-4}
        
        ERM \cite{ERM} & 0.94 (0.03) / 0.04 (0.01) / 0.54 (0.01) / 92.02 (0.35) & 0.95 (0.05) / 0.32 (0.04) / 0.67 (0.01) / \textbf{98.34 (0.17)} & 0.95 (0.03) / 0.15 (0.01) / 0.56 (0.01) / 97.99 (0.34) \\
        
        IRM \cite{IRM} & 0.96 (0.01) / 0.04 (0.01) / 0.53 (0.01) / 90.67 (0.89) & 0.95 (0.05) / 0.32 (0.03) / 0.67 (0.01) / 97.94 (0.25) & 0.95 (0.02) / 0.15 (0.03) / 0.55 (0.01) / 97.65 (0.28) \\
        
        GDRO \cite{GDRO} & 0.95 (0.01) / 0.04 (0.02) / 0.55 (0.01) / 93.00 (0.67) & 0.95 (0.02) / 0.31 (0.05) / 0.66 (0.01) / 98.07 (0.33) & 0.95 (0.04) / 0.16 (0.01) / 0.58 (0.01) / 97.84 (0.30) \\
        
        Mixup \cite{Mixup} & 0.95 (0.01) / 0.04 (0.01) / 0.54 (0.01) / 93.27 (0.84) & 0.95 (0.05) / 0.31 (0.05) / 0.66 (0.01) / 98.13 (0.20) & 0.95 (0.05) / 0.16 (0.02) / 0.57 (0.01) / \underline{98.26 (0.11)} \\
        
        MLDG \cite{MLDG} & 0.95 (0.01) / 0.04 (0.01) / 0.53 (0.01) / 92.37 (0.47) & 0.95 (0.03) / 0.31 (0.03) / 0.65 (0.01) / 97.65 (0.18) & 0.95 (0.05) / 0.16 (0.04) / 0.56 (0.01) / 98.07 (0.26) \\
        
        CORAL \cite{CORAL} & 0.95 (0.01) / 0.04 (0.01) / 0.55 (0.01) / 93.81 (0.82) & 0.96 (0.02) / 0.31 (0.03) / 0.67 (0.01) / \underline{98.31 (0.44)} & 0.96 (0.03) / 0.16 (0.05) / 0.58 (0.01) / \textbf{98.49 (0.29)} \\
        
        DANN \cite{DANN} & 0.94 (0.02) / 0.04 (0.01) / 0.54 (0.01) / 91.24 (2.11) & 0.93 (0.05) / 0.30 (0.02) / 0.63 (0.04) / 96.74 (0.27) & 0.93 (0.02) / \underline{0.14 (0.01)} / \underline{0.54 (0.03)} / 96.84 (0.34) \\
        
        CDANN \cite{CDANN} & 0.94 (0.01) / 0.04 (0.01) / 0.53 (0.01) / 91.08 (1.21) & 0.93 (0.05) / 0.31 (0.02) / 0.66 (0.01) / 97.47 (0.32) & 0.93 (0.01) / 0.15 (0.01) / 0.57 (0.02) / 96.57 (0.66) \\
        
        DDG \cite{DDG} & \textbf{0.97 (0.01)} / \textbf{0.01 (0.01)} / \textbf{0.50 (0.05)} / \textbf{96.90 (0.11)} & 0.96 (0.03) / 0.31 (0.04) / 0.65 (0.01) / 97.79 (0.05) & \underline{0.97 (0.02)} / 0.16 (0.01) / 0.59 (0.03) / 97.42 (0.33) \\

        \cmidrule(lr){1-4}

        DIR \cite{DIR} & 0.73 (0.03) / 0.02 (0.05) / 0.52 (0.05) / 71.89 (0.21) & 0.73 (0.03) / \underline{0.18 (0.03)} / 0.57 (0.05) / 72.61 (0.24) & 0.72 (0.02) / 0.17 (0.04) / 0.56 (0.01) / 71.72 (0.11) \\
        
        EIIL \cite{EIIL} & 0.93 (0.01) / 0.14 (0.04) / 0.58 (0.01) / 82.00 (0.76) & 0.96 (0.02) / 0.27 (0.03) / 0.63 (0.06) / 92.07 (0.18) & 0.96 (0.04) / \underline{0.14 (0.01)} / 0.61 (0.01) / 92.17 (0.28) \\
        
        FVAE \cite{FarconVAE} & 0.95 (0.02) / 0.07 (0.03) / 0.53 (0.03) / 91.44 (2.02) & \underline{0.96 (0.01)} / 0.30 (0.02) / 0.59 (0.06) / 92.49 (1.42) & 0.96 (0.06) / 0.18 (0.05) / 0.60 (0.04) / 91.69 (6.34) \\

        FATDM \cite{FATDM} & 0.94 (0.01) / \textbf{0.01 (0.01)} / 0.52 (0.02) / \underline{94.02 (1.02)} & 0.95 (0.01) / 0.19 (0.01) / \textbf{0.55 (0.02)} / 90.65 (1.42) & 0.94 (0.01) / 0.14 (0.02) / 0.55 (0.02) / 90.25 (1.36) \\
        
        \cmidrule(lr){1-4}

        \sysname{} & \underline{0.97 (0.02)} / \underline{0.02 (0.01)} / \underline{0.52 (0.01)} / 93.11 (1.23) & \textbf{0.99 (0.02)} / \textbf{0.18 (0.02)} / \underline{0.56 (0.04)} / 90.85 (1.56) & \textbf{0.99 (0.02)} / \textbf{0.12 (0.03)} / \textbf{0.54 (0.02)} / 91.77 (1.94) \\

        \bottomrule
    \end{tabular}
\end{table*}

\begin{table*}[t]
\scriptsize
    \centering
    \setlength{\abovecaptionskip}{1pt}
    \setlength{\belowcaptionskip}{0pt}
    \setlength\tabcolsep{3pt}
    \begin{tabular}{l|c|c|c|c}

        \toprule
          & $45^{\circ}$ & $60^{\circ}$ & $75^{\circ}$ & Avg \\
        \cmidrule(lr){1-1} \cmidrule(lr){2-2} \cmidrule(lr){3-3} \cmidrule(lr){4-4} \cmidrule(lr){5-5}

        ERM \cite{ERM} & 0.95 (0.04) / 0.35 (0.05) / 0.69 (0.01) / 98.34 (0.12) & 0.95 (0.01) / 0.29 (0.02) / 0.68 (0.01) / \underline{98.04 (0.18)} & 0.93 (0.01) / 0.17 (0.02) / 0.62 (0.02) / 94.60 (0.46) & 0.946 / 0.221 / 0.626 / 96.55\\
        
        IRM \cite{IRM} & 0.96 (0.05) / 0.35 (0.01) / 0.69 (0.01) / 97.68 (0.42) & 0.96 (0.01) / 0.28 (0.01) / 0.66 (0.01) / 97.11 (0.47) & 0.93 (0.02) / 0.16 (0.02) / 0.61 (0.01) / 93.67 (0.30) & 0.953 / 0.217 / 0.619 / 95.79 \\
        
        GDRO \cite{GDRO} & 0.95 (0.05) / 0.35 (0.01) / 0.71 (0.02) / 98.07 (0.30) & 0.96 (0.01) / 0.29 (0.01) / 0.69 (0.01) / 97.88 (0.39) & 0.93 (0.04) / 0.16 (0.01) / 0.61 (0.01) / 94.40 (0.41) & 0.952 / 0.220 / 0.631 / 96.54\\
        
        Mixup \cite{Mixup} & 0.95 (0.04) / 0.34 (0.03) / 0.69 (0.01) / \underline{98.39 (0.22)} & 0.96 (0.03) / 0.29 (0.04) / 0.68 (0.01) / 97.94 (0.14) & 0.93 (0.01) / 0.15 (0.01) / 0.59 (0.01) / 93.58 (0.61) & 0.951 / 0.215 / 0.623 / 96.59\\
        
        MLDG \cite{MLDG} & 0.95 (0.05) / 0.35 (0.01) / 0.70 (0.01) / 98.15 (0.07) & 0.96 (0.03) / 0.28 (0.04) / 0.66 (0.01) / 97.59 (0.15) & 0.94 (0.02) / 0.17 (0.04) / 0.62 (0.01) / 94.30 (0.36) & 0.952 / 0.219 / 0.620 / 96.36 \\
        
        CORAL \cite{CORAL} & 0.96 (0.05) / 0.35 (0.04) / 0.68 (0.01) / \textbf{98.63 (0.23)} & 0.96 (0.05) / 0.29 (0.03) / 0.68 (0.01) / \textbf{98.33 (0.16)} & 0.94 (0.01) / 0.16 (0.01) / 0.61 (0.02) / \underline{95.43 (0.74)} & 0.954 / 0.221 / 0.628 / \underline{97.17}\\
        
        DANN \cite{DANN} & 0.93 (0.02) / 0.35 (0.01) / 0.70 (0.01) / 97.36 (0.26) & 0.94 (0.04) / 0.29 (0.01) / 0.69 (0.01) / 97.03 (0.25) & 0.90 (0.01) / 0.17 (0.01) / 0.62 (0.01) / 90.60 (1.13) & 0.928 / 0.216 / 0.620 / 94.97\\
        
        CDANN \cite{CDANN} & 0.93 (0.03) / 0.35 (0.01) / 0.69 (0.02) / 97.61 (0.40) & 0.94 (0.02) / 0.29 (0.03) / 0.67 (0.01) / 97.60 (0.17) & 0.90 (0.02) / 0.18 (0.02) / 0.62 (0.01) / 90.63 (1.67) & 0.928 / 0.219 / 0.623 / 95.16\\
        
        DDG \cite{DDG} & 0.97 (0.02) / 0.35 (0.01) / 0.69 (0.05) / 97.97 (0.05) & 0.97 (0.03) / 0.28 (0.02) / 0.64 (0.05) / 97.81 (0.06) & 0.95 (0.03) / 0.15 (0.01) / 0.58 (0.01) / \textbf{96.74 (0.13)} & \underline{0.963} / 0.209 / 0.609 / \textbf{97.44}\\
         
        \cmidrule(lr){1-5}
        
       DIR \cite{DIR} & 0.73 (0.04) / \textbf{0.22 (0.02)} / 0.57 (0.03) / 72.35 (0.19) & 0.72 (0.04) / \textbf{0.21 (0.03)} / \textbf{0.56 (0.03)} / 70.85 (0.21) & 0.73 (0.02) / 0.16 (0.05) / \underline{0.57 (0.01)} / 69.69 (0.14) & 0.728 / \underline{0.161} / 0.555 / 71.52\\

        EIIL \cite{EIIL} & 0.97 (0.03) / 0.26 (0.02) / 0.62 (0.01) / 91.83 (0.38) & 0.96 (0.02) / 0.27 (0.01) / 0.59 (0.01) / 93.09 (0.22) & 0.96 (0.02) / 0.21 (0.02) / 0.61 (0.01) / 93.77 (0.10) & 0.959 / 0.216 / 0.607 / 90.82\\
       
        FVAE \cite{FarconVAE} & \underline{0.97 (0.01)} / 0.28 (0.04) / \textbf{0.56 (0.02)} / 92.85 (1.30) & \underline{0.97 (0.01)} / 0.28 (0.01) / 0.67 (0.03) / 91.02 (1.25) & 0.94 (0.02) / 0.21 (0.02) / 0.60 (0.03) / 91.34 (1.74) & 0.958 / 0.220 / 0.592 / 91.80 \\

        FATDM \cite{FATDM} & 0.96 (0.04) / \underline{0.25 (0.01)} / 0.57 (0.02) / 92.90 (1.21) & 0.95 (0.02) / 0.26 (0.03) / 0.57 (0.01) / 91.72 (1.32) & \underline{0.96 (0.01)} / \underline{0.14 (0.02)} / 0.57 (0.03) / 91.11 (0.84) & 0.953 / 0.165 / \underline{0.555} / 91.78\\
        
        \cmidrule(lr){1-5}

        \sysname{} & \textbf{0.98 (0.02)} / 0.28 (0.02) / \underline{0.56 (0.03)} / 92.05 (2.34) & \textbf{0.98 (0.02)} / \underline{0.24 (0.03)} / \underline{0.56 (0.04)} / 91.95 (2.23) & \textbf{0.98 (0.01)} / \textbf{0.11 (0.03)} / \textbf{0.56 (0.04)} / 91.55 (1.02) &  \textbf{0.980} / \textbf{0.157} / \textbf{0.552} / 91.88\\
        
        \bottomrule
    \end{tabular}
\end{table*}

\subsection{Datasets}
\label{sec:dataset}

\texttt{Rotated-Colored-MNIST}
(\texttt{RCMNIST}) dataset is a synthetic image dataset generated from the MNIST dataset \cite{lecun1998gradient} by rotating and coloring the digits. The rotation angles $d\in\ $\{$0^{\circ}$,$15^{\circ}$,$30^{\circ}$,$45^{\circ}$,$60^{\circ}$,$75^{\circ}$\} of the digits are used to partition different domains, while the color $a\in\{\text{red}, \text{green}\}$ of the digits is served as the sensitive attribute. A binary target label is created by grouping digits into $\{0,1,2,3,4\}$ and $\{5,6,7,8,9\}$. To investigate the robustness of \sysname{} in the face of correlation shift, we controlled the correlation between label and color for each domain in the generation process of \texttt{RCMNIST}, setting them respectively to $\{0, 0.8, 0.5, 0.1, 0.3, 0.6\}$. The correlation for domain $d=0^{\circ}$ was set to 0, implying that higher accuracy leads to fairer results.

\texttt{New-York-Stop-and-Frisk}
(\texttt{NYPD}) dataset \cite{goel2016precinct} is a real-world tabular dataset containing stop, question, and frisk data from some suspects in five different cities. We selected the full-year data from 2011, which had the highest number of stops compared to any other year. We consider the cities $d \in $\{BROOKLYN, QUEENS, MANHATTAN, BRONX, STATEN IS\} where suspects were sampled as domains. The suspects' gender $a \in \{\text{Male, Female}\}$ serves as the sensitive attribute, and whether a suspect was frisked is treated as the target label.

\texttt{FairFace}
dataset \cite{Fairface} is a novel face image dataset containing 108,501 images labeled with race, gender, and age groups which is balanced on race. The dataset comprises face images from seven race group $d \in$ \{White, Black, Latino/Hispanic, East Asian, Southeast Asian, Indian, Middle Eastern\}. These race groups determine the domain to which an image belongs. Gender $a \in \{\text{Male, Female}\}$ is considered a sensitive attribute, and the binary target label is determined based on whether the age is greater than 60 years old.

\subsection{Evaluation Metrics}

Given input feature $X\in\mathcal{X}$, target label $Y\in\mathcal{Y}=\{0,1\}$ and binary sensitive attribute $A\in\mathcal{A}=\{-1,1\}$, we evaluate the algorithm's performance on the test dataset $\mathcal{D}_{te}$. We measure the DG performance of the algorithm using \textbf{Accuracy} and evaluate the algorithm fairness using the following metrics.

\textbf{Demographic parity difference}
($\Delta_{DP}$) \cite{dwork2012fairness} is a type of \textit{group fairness} metric. Its rationale is that the acceptance rate provided by the algorithm should be the same across all sensitive subgroups. It can be formalized as
\begin{align*}
    \Delta_{DP}=\left|P(\hat{Y}=1|A=-1)-P(\hat{Y}=1|A=1)\right|,
\end{align*}
where $\hat{Y}$ is the predicted class label. The smaller the $\Delta_{DP}$, the fairer the algorithm.

\textbf{AUC for fairness} ($AUC_{fair}$) \cite{calders2013controlling} is a pairwise \textit{group fairness} metric. Define a scoring function $q_\theta : \mathcal{X} \to \mathbb{R}$, where $\theta$ represents the model parameters. The $AUC_{fair}$ of $q_\theta$ measures the probability of correctly ranking positive examples ahead of negative examples.
\begin{align*}  
    AUC_{fair}(q_\theta)=\frac{\sum_{X\in\mathcal{D}_{te}^{a=1}}\sum_{X'\in\mathcal{D}_{te}^{a=-1}}\mathds{1}[q_\theta(X)>q_\theta(X')]}{N^{a=1}\times N^{a=-1}},
 \end{align*}
where $\mathds{1}(\cdot)$ is an indicator function that returns 1 when the parameter is true and 0 otherwise. $\mathcal{D}_{te}$ is divided into $\mathcal{D}_{te}^{a=1}$ and $\mathcal{D}_{te}^{a=-1}$ based on $A$, which respectively contain $N^{a=1}$ and $N^{a=-1}$ samples. The value of $AUC_{fair}$ ranges from 0 to 1, with a value closer to 0.5 indicating a fairer algorithm.

\textbf{Consistency} \cite{LFR} is an \textit{individual fairness} metric based on the Lipschitz condition \cite{dwork2012fairness}. Specifically, \textit{Consistency} measures the distance between each individual and its \textit{k}-nearest neighbors. 
\begin{align*}
    Consistency=1 - \frac{1}{N}\sum_{i=1}^N \left|\hat{y}_i -\frac{1}{k} \sum_{j\in k\text{NN}(\textbf{x}_i)} \hat{y}_j\right|,
\end{align*}
where \textit{N} is the total number of samples in $\mathcal{D}_{te}$, $\hat{y}_i$ is the predicted class label for sample $\textbf{x}_i$, and $k\text{NN}(\cdot)$\footnote{Note that in \cite{LFR}, $k\text{NN}(\cdot)$ is applied to the full set of samples. To adapt it for DG task, here we apply it only to the set for the domain in which the samples are located.} takes the features of sample $\textbf{x}_i$ as input and returns the set of indices corresponding to its \textit{k}-nearest neighbors in the feature space. A larger \textit{Consistency} indicates a higher level of individual fairness.

\begin{table*}[t]
\scriptsize
    \centering
    \setlength{\abovecaptionskip}{0pt}
    \setlength{\belowcaptionskip}{2pt}
    \setlength\tabcolsep{4pt}
    \caption{Performance on \texttt{NYPD} (bold is the best, underline is the second best).}
    \label{tab:result-nypd}
    \begin{tabular}{l|c|c|c}
        \toprule
        \multirow{2}{*}{} & \multicolumn{3}{c}{Consisitency $\uparrow$ / $\Delta_{DP}$ $\downarrow$ / AUC$_{fair}$ $\downarrow$ / Accuracy $\uparrow$} \\ 
         \cmidrule(lr){2-4}
         & BROOKLYN & QUEENS & MANHATTAN \\
        \cmidrule(lr){1-1} \cmidrule(lr){2-2} \cmidrule(lr){3-3} \cmidrule(lr){4-4}

        ERM \cite{ERM} & 0.92 (0.03) / 0.14 (0.01) / 0.60 (0.03) / \textbf{62.57 (0.15)} & 0.92 (0.03) / 0.11 (0.01) / 0.58 (0.03) / 61.47 (0.15) & 0.91 (0.03) / 0.13 (0.04) / 0.60 (0.03) / 60.60 (0.16) \\
        
        IRM \cite{IRM} & 0.93 (0.02) / 0.17 (0.01) / 0.62 (0.05) / \underline{62.54 (0.07)} & 0.92 (0.03) / 0.13 (0.01) / 0.60 (0.01) / 61.80 (0.38) & 0.92 (0.01) / 0.15 (0.01) / 0.61 (0.01) / \underline{61.10 (0.13)} \\
        
        GDRO \cite{GDRO} & 0.93 (0.01) / 0.14 (0.01) / 0.60 (0.04) / 62.10 (0.17) & 0.92 (0.01) / 0.12 (0.01) / 0.59 (0.04) / 61.94 (0.30) & 0.92 (0.01) / 0.15 (0.01) / 0.60 (0.01) / 60.50 (0.07) \\
        
        Mixup \cite{Mixup} & 0.92 (0.03) / 0.13 (0.01) / 0.59 (0.01) / 62.24 (0.30) & 0.92 (0.01) / 0.10 (0.01) / 0.58 (0.01) / \underline{62.34 (0.98)} & 0.92 (0.01) / 0.13 (0.01) / 0.60 (0.01) / 60.17 (0.38) \\
        
        MLDG \cite{MLDG} & 0.93 (0.03) / 0.14 (0.01) / 0.60 (0.02) / 62.54 (0.13) & 0.92 (0.04) / 0.11 (0.01) / 0.58 (0.01) / 61.45 (0.23) & 0.92 (0.04) / 0.13 (0.05) / 0.60 (0.05) / 60.53 (0.18) \\
        
        CORAL \cite{CORAL} & 0.93 (0.02) / 0.15 (0.01) / 0.61 (0.01) / 62.38 (0.10) & 0.92 (0.01) / 0.11 (0.04) / 0.58 (0.01) / 61.51 (0.40) & 0.91 (0.02) / 0.13 (0.01) / 0.60 (0.01) / 60.61 (0.15) \\
        
        DANN \cite{DANN} & 0.92 (0.01) / 0.15 (0.02) / 0.61 (0.01) / 61.78 (0.32) & 0.92 (0.02) / 0.11 (0.01) / 0.58 (0.01) / 61.06 (1.33) & 0.91 (0.05) / 0.15 (0.02) / 0.60 (0.01) / 60.51 (0.57) \\
        
        CDANN \cite{CDANN} & 0.93 (0.05) / 0.15 (0.01) / 0.60 (0.01) / 62.07 (0.27) & 0.92 (0.02) / 0.11 (0.01) / 0.58 (0.01) / 61.28 (1.56) & 0.91 (0.04) / 0.15 (0.01) / 0.61 (0.01) / 60.59 (0.36) \\
        
        DDG \cite{DDG} & 0.94 (0.02) / 0.14 (0.01) / 0.60 (0.02) / 62.46 (0.11) & 0.94 (0.02) / 0.11 (0.01) / 0.58 (0.04) / \textbf{62.45 (0.13)} & 0.94 (0.03) / 0.13 (0.01) / 0.60 (0.04) / \textbf{61.11 (0.29)} \\

        \cmidrule(lr){1-4}
       
        DIR \cite{DIR} & 0.87 (0.03) / 0.14 (0.01) / 0.58 (0.05) / 57.23 (0.04) & 0.89 (0.01) / 0.10 (0.04) / 0.58 (0.05) / 55.80 (0.23) & 0.88 (0.02) / 0.11 (0.02) / 0.57 (0.02) / 56.19 (0.11) \\
        
        EIIL \cite{EIIL} & 0.94 (0.03) / 0.11 (0.01) / 0.59 (0.01) / 59.92 (1.16) & 0.94 (0.02) / 0.10 (0.01) / 0.58 (0.01) / 56.06 (0.24) & 0.93 (0.05) / \textbf{0.04 (0.01)} / \textbf{0.55 (0.01)} / 53.08 (0.98) \\
        
        FVAE \cite{FarconVAE} & \underline{0.95 (0.01)} / 0.12 (0.01) / 0.61 (0.01) / 58.78 (0.88) & \textbf{0.96 (0.02)} / 0.13 (0.01) / 0.58 (0.01) / 58.76 (3.17) & \underline{0.94 (0.01)} / 0.13 (0.01) / 0.61 (0.03) / 60.63 (2.95) \\
        
        FATDM \cite{FATDM} & 0.93 (0.01) / \textbf{0.09 (0.01)} / \underline{0.58 (0.02)} / 60.13 (1.10) & 0.93 (0.02) / \underline{0.05 (0.02)} / \textbf{0.56 (0.01)} / 58.48 (0.57) & \underline{0.94 (0.01)} / 0.12 (0.01) / 0.57 (0.01) / 57.02 (0.63) \\

        \cmidrule(lr){1-4}

        \sysname{} & \textbf{0.96 (0.01)} / \underline{0.10 (0.02)} / \textbf{0.58 (0.01)} / 58.08 (1.08) & \underline{0.96 (0.04)} / \textbf{0.03 (0.02)} / \underline{0.57 (0.01)} / 60.82 (0.55) & \textbf{0.95 (0.02)} / \underline{0.10 (0.01)} / \underline{0.56 (0.02)} / 58.14 (0.44) \\

        \bottomrule
    \end{tabular}
\end{table*}

\begin{table*}[t]
\scriptsize
    \centering
    \setlength{\abovecaptionskip}{-10pt}
    \setlength{\belowcaptionskip}{-10pt}
    \setlength\tabcolsep{4pt}
    \begin{tabular}{l|c|c|c}
        \toprule
         & BRONX & STATEN IS & Avg\\
        \cmidrule(lr){1-1} \cmidrule(lr){2-2} \cmidrule(lr){3-3} \cmidrule(lr){4-4}

        ERM \cite{ERM} & 0.90 (0.01) / 0.03 (0.03) / 0.55 (0.04) / 61.07 (0.46) & 0.91 (0.03) / 0.15 (0.01) / 0.61 (0.01) / \underline{67.02 (0.30)} & 0.910 / 0.113 / 0.588 / 62.55 \\
        
        IRM \cite{IRM} & 0.91 (0.04) / 0.06 (0.04) / 0.55 (0.02) / 59.84 (1.83) & 0.91 (0.01) / 0.17 (0.01) / 0.62 (0.01) / 66.68 (0.16) & 0.916 / 0.136 / 0.598 / 62.39 \\
        
        GDRO \cite{GDRO} & 0.91 (0.04) / 0.04 (0.03) / 0.53 (0.02) / 60.94 (1.73) & 0.91 (0.04) / 0.15 (0.01) / 0.60 (0.01) / 66.48 (0.20) & 0.914 / 0.121 / 0.585 / 62.39 \\
        
        Mixup \cite{Mixup} & 0.90 (0.02) / 0.07 (0.02) / 0.56 (0.01) / 61.30 (1.96) & 0.91 (0.02) / 0.14 (0.01) / 0.59 (0.05) / 66.25 (0.85) & 0.914 / 0.113 / 0.583 / 62.46 \\
        
        MLDG \cite{MLDG} & 0.91 (0.05) / 0.03 (0.01) / 0.53 (0.02) / 60.94 (2.43) & 0.91 (0.04) / 0.15 (0.04) / 0.61 (0.03) / 66.94 (0.25) & 0.916 / 0.113 / 0.585 / 62.48 \\
        
        CORAL \cite{CORAL} & 0.91 (0.01) / 0.04 (0.03) / 0.54 (0.02) / 61.52 (3.13) & 0.91 (0.02) / 0.15 (0.01) / 0.60 (0.02) / \textbf{67.08 (0.21)} & 0.917 / 0.114 / 0.586 / \underline{62.62} \\
        
        DANN \cite{DANN} & 0.88 (0.03) / 0.10 (0.01) / 0.56 (0.02) / 58.32 (1.28) & 0.91 (0.03) / 0.14 (0.01) / 0.60 (0.01) / 65.62 (0.18) & 0.910 / 0.130 / 0.591 / 61.46 \\
        
        CDANN \cite{CDANN} & 0.90 (0.02) / 0.09 (0.03) / 0.56 (0.02) / 61.26 (1.25) & 0.91 (0.05) / 0.17 (0.01) / 0.61 (0.01) / 66.07 (0.59) & 0.914 / 0.132 / 0.594 / 62.25 \\
        
        DDG \cite{DDG} & 0.93 (0.03) / 0.02 (0.02) / \underline{0.53 (0.01)} / \textbf{64.91 (0.57)} & 0.93 (0.04) / 0.15 (0.01) / 0.60 (0.01) / 66.46 (0.22) & 0.935 / 0.109 / 0.582 / \textbf{63.48} \\
          
        \cmidrule(lr){1-4}
        
        DIR \cite{DIR} & 0.90 (0.04) / 0.08 (0.02) / 0.58 (0.03) / 54.25 (0.17) & 0.89 (0.03) / \textbf{0.11 (0.02)} / 0.56 (0.01) / 55.19 (0.11) & 0.883 / 0.107 / 0.577 / 55.73 \\
        
        EIIL \cite{EIIL} & 0.92 (0.04) / 0.03 (0.02) / \underline{0.53 (0.01)} / 61.02 (1.14) & \underline{0.94 (0.02)} / 0.13 (0.01) / \textbf{0.55 (0.01)} / 56.69 (0.98) & 0.933 / \underline{0.080} / \underline{0.561} / 57.35 \\

        FVAE \cite{FarconVAE} & 0.93 (0.04) / 0.04 (0.01) / 0.54 (0.01) / 61.08 (1.16) & 0.93 (0.02) / 0.16 (0.01) / 0.56 (0.03) / 63.96 (1.58) & \underline{0.941} / 0.115 / 0.578 / 60.64 \\

        FATDM \cite{FATDM} & \underline{0.94 (0.02)} / \textbf{0.01 (0.02)} / 0.54 (0.01) / 62.57 (0.59) & 0.93 (0.05) / 0.14 (0.01) / 0.57 (0.02) / 62.80 (1.83) & 0.931 / 0.082 / 0.566 / 60.20 \\
        
        \cmidrule(lr){1-4}

        \sysname{} & \textbf{0.94 (0.01)} / \underline{0.02 (0.02)} / \textbf{0.52 (0.01)} / \underline{63.87 (1.14)} & \textbf{0.95 (0.05)} / \underline{0.12 (0.01)} / \underline{0.55 (0.03)} / 62.63 (1.06) & \textbf{0.955} / \textbf{0.073} / \textbf{0.560} / 60.71 \\

        \bottomrule
    \end{tabular}
\end{table*}

\subsection{Compared Methods}

We validate the utility of \sysname{} in handling Problem \ref{prob:problem} using 13 methods. ERM \cite{ERM}, IRM \cite{IRM}, GDRO \cite{GDRO}, Mixup \cite{Mixup}, MLDG \cite{MLDG}, CORAL \cite{CORAL}, DANN \cite{DANN}, CDANN \cite{CDANN}, and DDG \cite{DDG} are DG methods without fairness consideration. Among them, DDG is a recently proposed method that focuses on learning invariant representations through disentanglement. DIR \cite{DIR} is a classic group fairness algorithm. EIIL \cite{EIIL} and FVAE \cite{FarconVAE} can achieve both domain generalization under correlation shift and fairness. FATDM \cite{FATDM} is the latest work that explicitly focuses on both domain generalization under covariate shift and group fairness simultaneously.




\section{Results}
\label{sec:res}
To evaluate the performance of \sysname{}, we posed the following research questions from shallow to deep and answered them in Sections \ref{sec:overall-res}, \ref{sec:trade-off} and \ref{sec:ablation}.
\begin{itemize}
    \item \textit{Q1)} Can \sysname{} effectively address Problem \ref{prob:problem}, or in other words, can \sysname{} ensure both group fairness and individual fairness on unseen domains while maximizing DG performance?
    \item \textit{Q2)} Does \sysname{} exhibit a good trade-off between DG performance and fairness?
    \item \textit{Q3)} What are the roles of the transformation model $T$ and the fair representation learner $g$ in \sysname{}?
    \item \textit{Q4)} How is $R_{fair}$ ensuring algorithmic fairness in the learning process of \sysname{}?
\end{itemize}




\subsection{Overall Performance}
\label{sec:overall-res}

The overall performance of \sysname{} and its competing methods on three real-world datasets is presented in Table \ref{tab:result-rcmnist}, \ref{tab:result-nypd} and \ref{tab:result-fairface}, $\uparrow$ means higher is better, $\downarrow$ means lower is better. Each experiment was conducted five times and the average results were recorded, with standard deviations reported in parentheses.

\textbf{Fairness Evaluation.} Focus on the average of each fairness metric across all domains, \sysname{} almost achieves the best performance on all three datasets. Excluding DIR, which is not competitive due to its poor DG performance, \sysname{} consistently ranks as either the fairest or the second fairest in each domain. This indicates its relative stability in achieving fairness across various domains compared to competing methods. All of the above analyses shows that \sysname{} is able to achieve both individual fairness and group fairness on unseen domains with state-of-the-art results.

\textbf{DG Evaluation Considering Trade-off.} Considering the accur- acy-fairness trade-off, we aim to enhance DG performance while simultaneously ensuring algorithmic fairness. From this perspective, we notice that (i) methods solely focusing on DG cannot ensure algorithmic fairness effectively. (ii) Although lower than the above methods, the performance of DG for \sysname{} is still competitive, and it outperforms other competing algorithms that also focus on fairness. (iii) On the \texttt{FairFace} dataset, \sysname{} ensures the best fairness while its DG performance is second only to DDG. This is because the transformation model allows \sysname{} to learn better domain-invariant representations when dealing with relatively complex data (facial photos) and types of environments.

Overall, \sysname{} ensures fairness on both tabular and image data while maintaining strong DG capabilities. It can learn a fairness-aware domain-invariant predictor to effectively address Problem \ref{prob:problem}. The success of \sysname{} on all three datasets, particularly \texttt{RCMNIST}, also demonstrates that our approach works effectively when dealing with DG problems involving covariate shift and correlation shift.

\begin{figure}[ht]
    \centering
    \setlength{\abovecaptionskip}{1pt}
    \setlength{\belowcaptionskip}{-10pt}
    \begin{subfigure}[b]{0.48\textwidth}
        \centering
        \setlength{\abovecaptionskip}{1pt}
    \setlength{\belowcaptionskip}{0pt}
        \includegraphics[width=\textwidth, trim=0 1.2cm 0 0, clip]{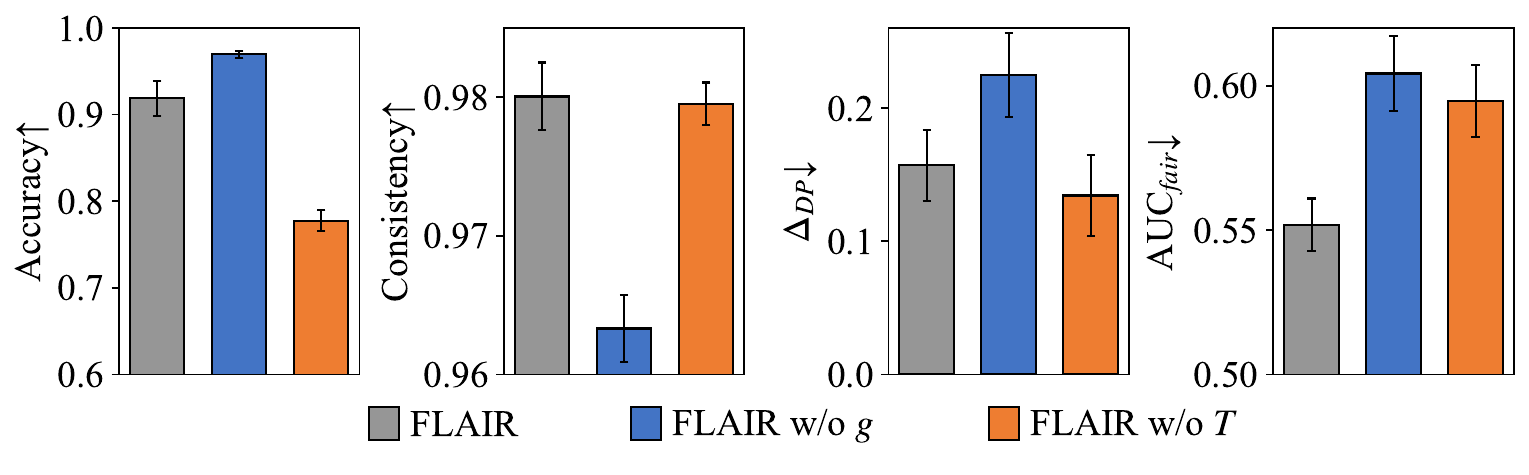}
    \caption{}
    \label{fig:ablation-rcmnist}
    \end{subfigure}
    \hfill
    \begin{subfigure}[b]{0.48\textwidth}
        \centering
        \setlength{\abovecaptionskip}{1pt}
    \setlength{\belowcaptionskip}{0pt}
        \includegraphics[width=\textwidth, trim=0 1.2cm 0 0, clip]{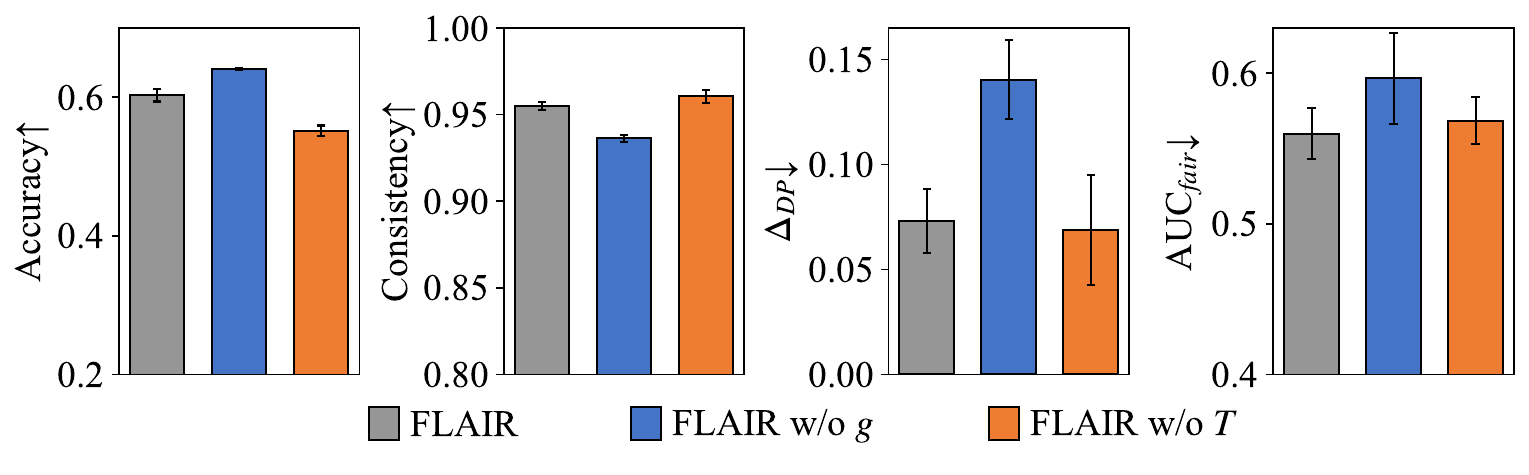}
            \caption{}
            \label{fig:ablation-nypd}
    \end{subfigure}
    \hfill
    \begin{subfigure}[b]{0.48\textwidth}
        \centering
        \setlength{\abovecaptionskip}{1pt}
    \setlength{\belowcaptionskip}{0pt}
        \includegraphics[width=\textwidth]{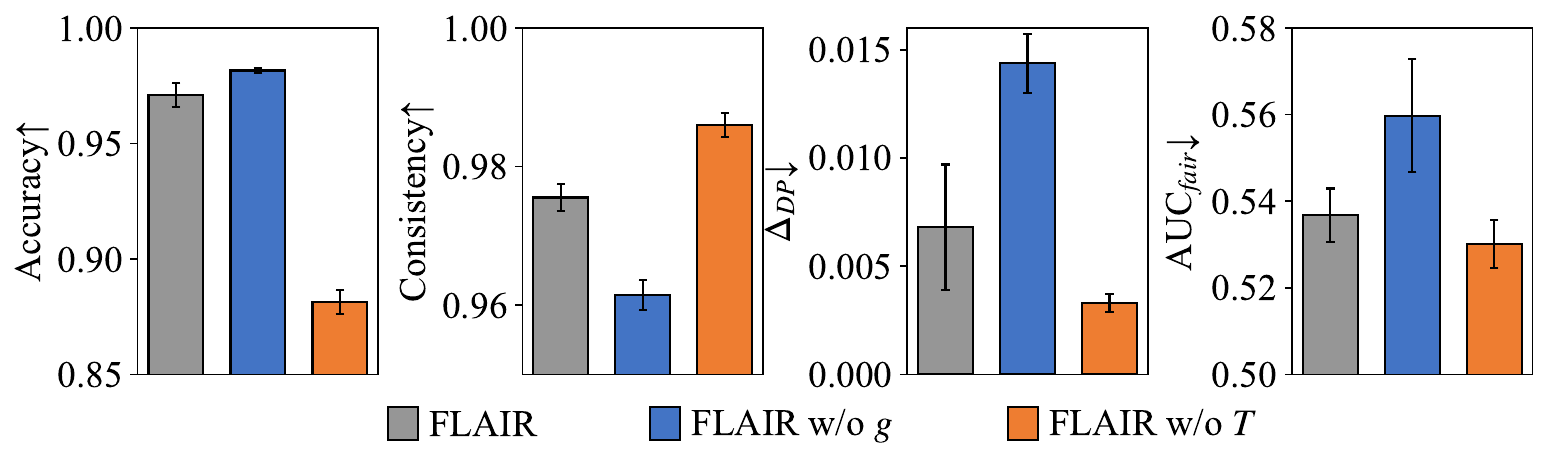}
            \caption{ }
            \label{fig:ablation}
    \end{subfigure}
    \caption{Ablation study over four metrics for \sysname{} and its two variants on (a) \texttt{RCMNIST}, (b) \texttt{NYPD} and (c) \texttt{FairFace} datasets. Results are averaged across all domains.}
    \label{fig:ablation_study}
\end{figure}

\begin{table*}[t]
\scriptsize
    \centering
    \setlength{\abovecaptionskip}{0pt}
    \setlength{\belowcaptionskip}{2pt}
    \setlength\tabcolsep{0.5pt}
    \caption{Performance on \texttt{FairFace} (bold is the best, underline is the second best).}
    \label{tab:result-fairface}
    \begin{tabular}{l|c|c|c|c}
        \toprule
        \multirow{2}{*}{} & \multicolumn{4}{c}{Consisitency $\uparrow$ / $\Delta_{DP}$ $\downarrow$ / AUC$_{fair}$ $\downarrow$ / Accuracy $\uparrow$} \\
         \cmidrule(lr){2-5}
         & White & Black & Latino/Hispanic & East Asian \\
        \cmidrule(lr){1-1} \cmidrule(lr){2-2} \cmidrule(lr){3-3} \cmidrule(lr){4-4} \cmidrule(lr){5-5}

        ERM \cite{ERM} & 0.95 (0.02) / 0.05 (0.02) / 0.57 (0.01) / 92.97 (3.04) & 0.95 (0.01) / 0.03 (0.01) / 0.58 (0.01) / 91.31 (2.28) & 0.96 (0.04) / 0.02 (0.01) / 0.57 (0.02) / 95.33 (1.51) & 0.96 (0.03) / 0.02 (0.01) / 0.60 (0.01) / 96.71 (0.28) \\
        IRM \cite{IRM} & 0.95 (0.01) / 0.05 (0.01) / 0.57 (0.01) / 93.26 (1.17) & 0.96 (0.01) / 0.02 (0.01) / 0.57 (0.01) / 92.94 (1.01) & 0.96 (0.03) / 0.03 (0.02) / 0.56 (0.03) / 94.85 (2.27) & 0.95 (0.01) / 0.03 (0.01) / 0.60 (0.01) / 95.23 (1.26) \\
        GDRO \cite{GDRO} & 0.95 (0.01) / 0.06 (0.01) / 0.57 (0.01) / 91.94 (0.47) & 0.96 (0.01) / 0.02 (0.01) / 0.58 (0.01) / 92.67 (1.59) & 0.96 (0.01) / 0.01 (0.02) / 0.57 (0.01) / 96.14 (0.69) & 0.96 (0.01) / 0.02 (0.01) / 0.61 (0.01) / 96.63 (0.30) \\
        Mixup \cite{Mixup} & 0.95 (0.04) / 0.05 (0.01) / 0.55 (0.05) / 92.29 (2.26) & 0.95 (0.01) / 0.03 (0.01) / 0.53 (0.01) / 92.49 (2.20) & 0.96 (0.03) / 0.02 (0.03) / 0.54 (0.03) / 95.42 (1.02) & 0.96 (0.01) / 0.03 (0.01) / 0.55 (0.02) / 96.09 (1.43) \\
        MLDG \cite{MLDG} & 0.95 (0.05) / 0.05 (0.01) / 0.58 (0.01) / 93.71 (0.41) & 0.95 (0.04) / 0.02 (0.01) / 0.58 (0.01) / 93.21 (0.40) & 0.96 (0.03) / 0.02 (0.02) / 0.58 (0.01) / 95.64 (0.93) & 0.95 (0.03) / 0.02 (0.01) / 0.59 (0.02) / 96.34 (0.84) \\
        CORAL \cite{CORAL} & 0.95 (0.05) / 0.07 (0.02) / 0.57 (0.01) / 91.99 (2.14) & 0.95 (0.04) / 0.05 (0.01) / 0.57 (0.01) / 89.44 (2.35) & 0.95 (0.05) / 0.04 (0.01) / 0.56 (0.01) / 93.97 (0.57) & 0.95 (0.02) / 0.04 (0.01) / 0.57 (0.03) / 94.84 (0.99) \\
        DANN \cite{DANN} & 0.93 (0.04) / 0.11 (0.01) / 0.59 (0.01) / 80.72 (2.23) & 0.92 (0.01) / 0.09 (0.01) / 0.58 (0.01) / 74.07 (1.65) & 0.93 (0.04) / 0.09 (0.02) / 0.61 (0.04) / 87.17 (2.48) & 0.92 (0.02) / 0.10 (0.02) / 0.60 (0.01) / 85.42 (3.67) \\
        CDANN \cite{CDANN} & 0.91 (0.05) / 0.12 (0.02) / 0.59 (0.01) / 76.13 (2.65) & 0.91 (0.03) / 0.08 (0.01) / 0.58 (0.01) / 76.43 (2.09) & 0.92 (0.01) / 0.06 (0.01) / 0.60 (0.01) / 91.03 (1.42) & 0.92 (0.02) / 0.08 (0.02) / 0.60 (0.01) / 89.09 (2.69) \\
        DDG \cite{DDG} & 0.95 (0.04) / 0.04 (0.04) / 0.56 (0.01) / \underline{96.25 (0.64)} & 0.96 (0.05) / 0.03 (0.01) / 0.55 (0.01) / \underline{97.26 (0.60)} & 0.97 (0.03) / \underline{0.01 (0.01)} / 0.55 (0.02) / \underline{98.15 (0.52)} & 0.96 (0.03) / 0.02 (0.01) / 0.59 (0.01) / \textbf{98.37 (0.63)} \\
        
        \cmidrule(lr){1-5}
        
       DIR \cite{DIR} & 0.74 (0.04) / 0.02 (0.03) / \textbf{0.52 (0.05)} / 76.14 (0.11) & 0.75 (0.04) / 0.03 (0.01) / \textbf{0.52 (0.03)} / 76.31 (0.17) & 0.75 (0.04) / 0.03 (0.03) / \textbf{0.52 (0.02)} / 76.65 (0.19) & 0.75 (0.01) / 0.03 (0.01) / \textbf{0.52 (0.04)} / 77.42 (0.21) \\
        
        EIIL \cite{EIIL} & \underline{0.96 (0.01)} / \textbf{0.01 (0.00)} / 0.55 (0.01) / 89.92 (0.12) & \underline{0.96 (0.05)} / \textbf{0.00 (0.00)} / 0.59 (0.01) / 96.79 (0.11) & 0.97 (0.05) / 0.11 (0.03) / 0.55 (0.01) / 83.76 (0.57) & 0.97 (0.05) / 0.07 (0.01) / 0.60 (0.03) / 86.35 (1.87) \\

        FVAE \cite{FarconVAE} & 0.94 (0.01) / 0.05 (0.04) / 0.54 (0.02) / 90.36 (1.05) & 0.91 (0.02) / 0.03 (0.01) / 0.57 (0.04) / 89.63 (2.22) & \underline{0.97 (0.02)} / 0.03 (0.01) / 0.55 (0.01) / 93.30 (0.97) & \textbf{0.98 (0.01)} / 0.05 (0.01) / 0.57 (0.02) / 91.44 (1.58) \\

        FATDM \cite{FATDM} & 0.95 (0.01) / 0.02 (0.02) / \underline{0.53 (0.03)} / 96.23 (1.11) & 0.96 (0.01) / 0.02 (0.02) / 0.54 (0.01) / 95.82 (0.13) & 0.95 (0.01) / 0.02 (0.04) / 0.55 (0.02) / 95.38 (0.29) & 0.95 (0.03) / \underline{0.01 (0.01)} / 0.55 (0.04) / 96.31 (0.35) \\

        \cmidrule(lr){1-5}

        \sysname{} & \textbf{0.98 (0.01)} / \underline{0.02 (0.01)} / 0.57 (0.01) / \textbf{96.56 (0.76)} & \textbf{0.98 (0.01)} / \underline{0.01 (0.00)} / \underline{0.53 (0.01)} / \textbf{97.60 (0.13)} & \textbf{0.98 (0.01)} / \textbf{0.00 (0.00)} / \underline{0.54 (0.01)} / \textbf{98.31 (0.30)} & \underline{0.97 (0.02)} / \textbf{0.00 (0.00)} / \underline{0.55 (0.01)} / \underline{97.36 (0.28)} \\
        
        \bottomrule
    \end{tabular}
\end{table*}

\begin{table*}[t]
\scriptsize
    \centering
    \setlength{\abovecaptionskip}{1pt}
    \setlength{\belowcaptionskip}{0pt}
    \setlength\tabcolsep{1pt}
    \begin{tabular}{l|c|c|c|c}
        \toprule
         & Southeast Asian & Indian & Middle Eastern & Avg \\
        \cmidrule(lr){1-1} \cmidrule(lr){2-2} \cmidrule(lr){3-3} \cmidrule(lr){4-4} \cmidrule(lr){5-5}

        ERM \cite{ERM} & 0.96 (0.03) / 0.01 (0.01) / 0.56 (0.01) / 94.42 (0.29) & 0.94 (0.04) / 0.01 (0.03) / 0.53 (0.02) / 94.66 (0.32) & 0.95 (0.02) / 0.04 (0.01) / 0.57 (0.02) / 93.42 (1.85) & 0.952 / 0.026 / 0.568 / 94.12\\
        
        IRM \cite{IRM} & 0.96 (0.05) / 0.01 (0.01) / 0.56 (0.01) / 94.56 (0.41) & 0.93 (0.04) / 0.02 (0.02) / 0.54 (0.02) / 94.27 (0.29) & 0.94 (0.04) / 0.04 (0.01) / 0.56 (0.01) / 93.95 (1.37) & 0.950 / 0.029 / 0.566 / 94.15\\

        GDRO \cite{GDRO} & 0.96 (0.01) / 0.02 (0.01) / 0.57 (0.02) / 94.26 (0.47) & 0.94 (0.02) / 0.02 (0.01) / 0.53 (0.01) / 93.99 (0.79) & 0.95 (0.03) / 0.04 (0.01) / 0.57 (0.01) / 93.87 (0.47) & 0.954 / 0.027 / 0.570 / 94.21 \\
        
        Mixup \cite{Mixup} & 0.96 (0.04) / 0.01 (0.02) / \underline{0.51 (0.01)} / 94.55 (0.28) & 0.94 (0.01) / 0.03 (0.01) / 0.53 (0.01) / 93.76 (0.46) & 0.95 (0.05) / 0.04 (0.01) / 0.55 (0.02) / 93.83 (0.44) & 0.953 / 0.029 / 0.538 / 94.06\\
        
        MLDG \cite{MLDG} & 0.96 (0.05) / \underline{0.01 (0.02)} / 0.56 (0.02) / 94.62 (0.16) & 0.93 (0.03) / 0.02 (0.02) / 0.56 (0.02) / 94.68 (0.37) & 0.95 (0.03) / 0.03 (0.01) / 0.58 (0.01) / 94.57 (0.20) & 0.952 / 0.023 / 0.577 / 94.68\\
        
        CORAL \cite{CORAL} & 0.96 (0.05) / 0.02 (0.01) / 0.54 (0.02) / 93.96 (0.74) & 0.93 (0.05) / 0.03 (0.01) / 0.54 (0.02) / 93.78 (0.45) & 0.94 (0.02) / 0.05 (0.01) / 0.56 (0.02) / 92.56 (0.73) & 0.949 / 0.043 / 0.558 / 92.93\\
        
        DANN \cite{DANN} & 0.91 (0.05) / 0.04 (0.01) / 0.56 (0.01) / 86.96 (1.54) & 0.90 (0.04) / 0.07 (0.01) / 0.58 (0.02) / 88.35 (1.85) & 0.92 (0.02) / 0.09 (0.02) / 0.60 (0.02) / 84.68 (3.25) & 0.918 / 0.082 / 0.590 / 83.91\\
        
        CDANN \cite{CDANN} & 0.93 (0.05) / 0.04 (0.02) / 0.55 (0.01) / 84.56 (2.98) & 0.91 (0.02) / 0.06 (0.03) / 0.56 (0.03) / 88.91 (3.54) & 0.93 (0.02) / 0.05 (0.04) / 0.58 (0.01) / 86.14 (5.08) & 0.918 / 0.070 / 0.581 / 84.61\\
        
        DDG \cite{DDG} & 0.97 (0.03) / 0.01 (0.02) / 0.54 (0.01) / \textbf{97.98 (0.21)} & 0.94 (0.04) / \underline{0.01 (0.01)} / 0.54 (0.02) / \textbf{97.29 (0.46)} & 0.95 (0.01) / 0.04 (0.04) / 0.55 (0.02) / \textbf{97.13 (0.68)} & 0.959 / 0.023 / 0.554 / \textbf{97.49}\\
    
        \cmidrule(lr){1-5}

        DIR \cite{DIR} & 0.75 (0.03) / 0.03 (0.04) / 0.52 (0.02) / 75.46 (0.20) & 0.74 (0.05) / 0.03 (0.01) / \textbf{0.52 (0.03)} / 74.55 (0.31) & 0.75 (0.01) / 0.03 (0.03) / \textbf{0.52 (0.05)} / 68.14 (4.08) & 0.748 / 0.027 / \textbf{0.521} / 74.95\\
        
        EIIL \cite{EIIL} & \underline{0.97 (0.01)} / 0.03 (0.01) / 0.54 (0.03) / 85.90 (0.82) & \underline{0.96 (0.02)} / 0.04 (0.01) / 0.55 (0.01) / 88.96 (0.57) & 0.96 (0.02) / 0.04 (0.02) / 0.56 (0.01) / 89.65 (0.26) & \underline{0.966} / 0.044 / 0.561 / 88.76\\
        
        FVAE \cite{FarconVAE} & 0.95 (0.01) / 0.03 (0.01) / 0.52 (0.01) / 90.23 (1.43) & 0.96 (0.04) / 0.04 (0.01) / 0.54 (0.01) / 88.48 (1.18) & \underline{0.96 (0.01)} / 0.06 (0.01) / 0.55 (0.02) / 86.80 (2.15) & 0.954 / 0.041 / 0.550 / 90.04\\

        FATDM \cite{FATDM} & 0.95 (0.01) / 0.01 (0.01) / 0.53 (0.02) / 94.21 (1.45) & 0.95 (0.01) / 0.01 (0.05) / 0.54 (0.03) / 94.52 (1.09) & 0.95 (0.05) / \underline{0.02 (0.01)} / \underline{0.54 (0.01)} / 94.01 (0.58) & 0.954 / \underline{0.017} / 0.539 / 95.21 \\

        \cmidrule(lr){1-5}

        \sysname{} & \textbf{0.98 (0.01)} / \textbf{0.00 (0.00)} / \textbf{0.51 (0.01)} / \underline{96.75 (1.12)} & \textbf{0.98 (0.01)} / \textbf{0.00 (0.00)} / \underline{0.53 (0.01)} / \underline{96.87 (0.12)} & \textbf{0.97 (0.02)} / \textbf{0.02 (0.00)} / \underline{0.54 (0.01)} / \underline{96.28 (0.89)} & \textbf{0.976} / \textbf{0.007} / \underline{0.537} / \underline{97.10}\\
        
        \bottomrule
    \end{tabular}
\end{table*}

\begin{figure*}
	\centering
    \setlength{\abovecaptionskip}{0pt}
    \setlength{\belowcaptionskip}{-10pt}
    \includegraphics[width=0.95\linewidth]{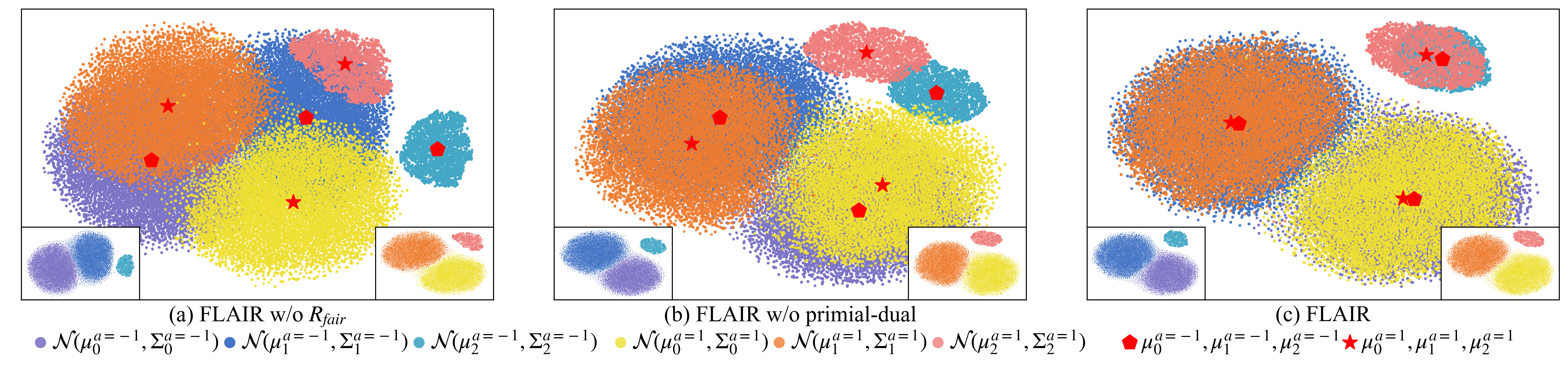}
    \caption{t-SNE visualization of the representations learned by (c) \sysname{} and its variants (a) \sysname{} w/o $R_{fair}$ and (b) \sysname{} w/o primal-dual on \texttt{RCMNIST} dataset. The main parts of (a)-(c) simultaneously visualize representations of two sensitive subgroups in the same latent space $\mathcal{C}$, while the bottom-left ($a=-1$) and bottom-right ($a=1$) visualize each group separately.}
    \label{fig:tsne}
\end{figure*}

\subsection{Ablation Study}
\label{sec:ablation}


To understand the roles of the transformation model $T$ and the fair representation learner $g$ in learning a fairness-aware domain invariant predictor, we constructed two different variants of \sysname{} for experimentation. They are: (i) \sysname{} \textit{w/o g}: remove $g$, i.e., learn a predictor $f_{v1} = h_s \circ \omega$. (ii) \sysname{} \textit{w/o T}: replace $T$ with a standard featurizer $h: \mathcal{X} \rightarrow \mathcal{X}^{'} \subseteq\mathbb{R}^{d^{'}}$ and modify the corresponding input and output dimensions of $g$ and $\omega$, i.e., learn a predictor $f_{v2} = h \circ g \circ \omega.$ The results of ablation study for \sysname{} and its two variants on three dataset are shown in Figure \ref{fig:ablation_study} (a), (b) and (c).


By comparing \sysname{} with its variant \sysname{} w/o $g$, we can see that the representations obtained by $T$ exhibit strong domain invariance but do not ensure fairness. Additionally, the improvement of \sysname{} on all three fairness metrics suggests that $g$ can simultaneously enhance individual and group fairness. The difference between the results of them further validates the accuracy-fairness trade-off .

Contrasting \sysname{} with its variant \sysname{} w/o $T$ further highlights the DG utility of $T$. At the same time, it's evident that while $g$ focuses only on fairness, it doesn't necessarily result in fairer outcomes. The reason for this is that the fair representation obtained solely through $g$ lacks domain invariance. As a result, it cannot handle covariate shift and correlation shift when generalizing to unseen domains. 

\textbf{The Utility of $R_{fair}$} To understand how the critical component $R_{fair}$ in $g$ promotes algorithmic fairness, we created two new variants of \sysname. They are (i) \sysname{} $w/o\ R_{fair}$: removing $R_{fair}$ from $g$ and (ii) \sysname{} $w/o\ primal$-$dual$: replacing the primal-dual updates with fixed parameters $\lambda_2$. Figure \ref{fig:tsne} shows the visualization of the fair content representations $\{\Tilde{\mathbf{c}}_i^a\}_{i=1}^{N^a}$ obtained by $g$ and its two variants on \texttt{RCMNIST}, organized by the respective sensitive subgroups.

The transition from (a) to (b) and (c) clearly shows that during optimization $R_{fair}$ brings the representations of the two sensitive subgroups closer in the latent space, ensuring that similar individuals from different groups get more similar representations. Additionally, the clustering of each sensitive subgroup can bring closer the distances between similar individuals within the same group. Combining above two points, $R_{fair}$ enables \sysname{} to achieve a strong individual fairness effect. At the same time, $R_{fair}$ enforces statistical parity between sensitive subgroups, reducing the distances between corresponding prototypes of different groups. This also ensures that \sysname{} achieves group fairness. The transition from (b) to (c) shows that optimizing through the primal-dual algorithm is able to achieve better algorithmic fairness performance.

\begin{figure}[t]
    \centering
    \setlength{\abovecaptionskip}{1pt}
    \setlength{\belowcaptionskip}{-10pt}
    \includegraphics[width=0.8\linewidth]{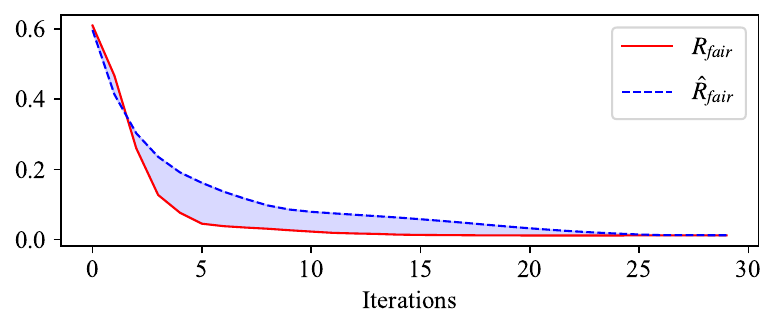}
    \caption{The convergence curves of $R_{fair}$ and $\hat{R}_{fair}$ during training. Both of them converge after 30 iterations.}
    \label{fig:convergence}
\end{figure}

The convergence curves for both $R_{fair}$ and $\hat{R}_{fair}$ during training are shown in Figure \ref{fig:convergence}. Since the prior $\pi$ updates are not fully synchronized with the posterior $\gamma$ updates (as seen in line 19 of Algorithm \ref{alg:algo}), a gap (indicated by the light blue area) exists between the two curves. However, their convergence trends are consistent, indicating that during training, $\hat{R}_{fair}$ can successfully approximate $R_{fair}$ and does not affect the successful convergence of $\mathcal{L}_{gmm}$.

\begin{figure}
		\centering
            \setlength{\abovecaptionskip}{6pt}
            \setlength{\belowcaptionskip}{-10pt}
            \includegraphics[width=\linewidth]{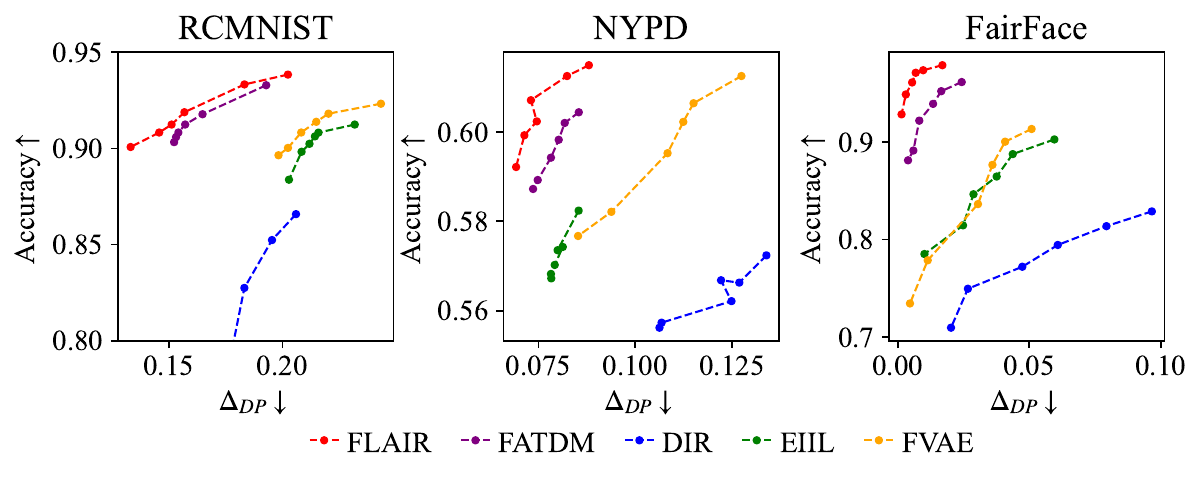}
    \caption{Accuracy-fairness trade-off across different methods by various $\lambda_2 \in \{0.05, 0.1, 0.5, 1, 2, 5\}$. The upper left indicates a better trade-off.}
    \label{fig:trade-off}
\end{figure}

\begin{figure}[ht]
    \centering
    \setlength{\abovecaptionskip}{1pt}
    \setlength{\belowcaptionskip}{-13pt}
    \begin{subfigure}[b]{0.48\textwidth}
        \centering
        \setlength{\abovecaptionskip}{1pt}
    \setlength{\belowcaptionskip}{0pt}
        \includegraphics[width=\textwidth]{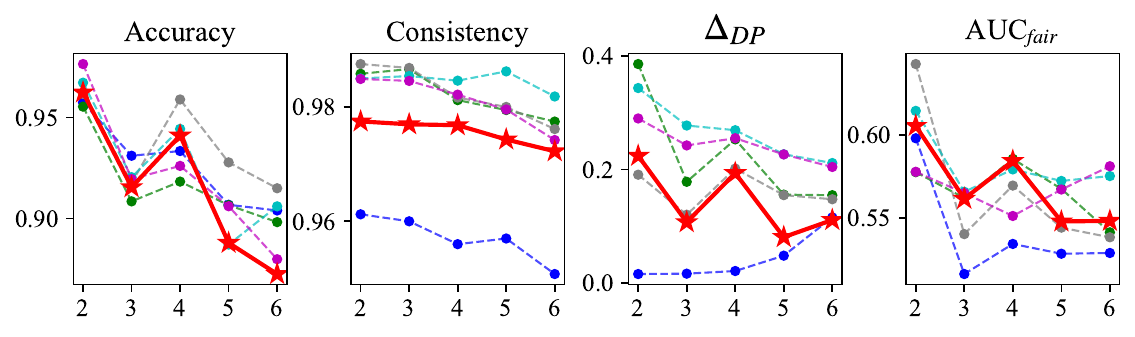}
    \caption{}
    \label{fig:nGMMs_mnist}
    \end{subfigure}
    \hfill
    \begin{subfigure}[b]{0.48\textwidth}
        \centering
        \setlength{\abovecaptionskip}{1pt}
    \setlength{\belowcaptionskip}{0pt}
        \includegraphics[width=\textwidth]{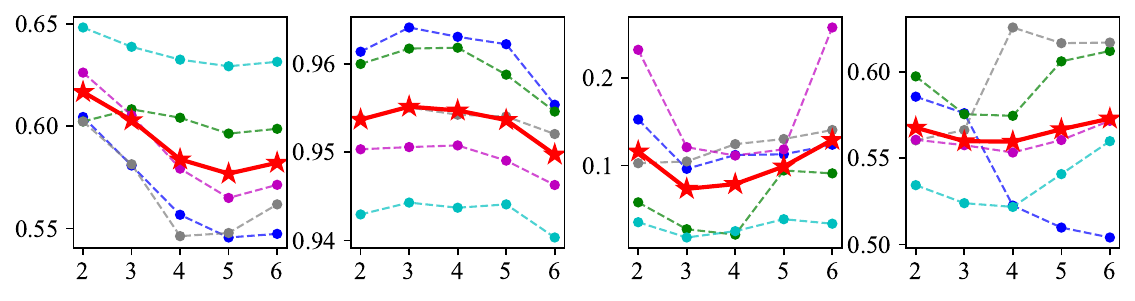}
            \caption{}
            \label{fig:nGMMs_nypd}
    \end{subfigure}
    \hfill
    \begin{subfigure}[b]{0.48\textwidth}
        \centering
        \setlength{\abovecaptionskip}{1pt}
    \setlength{\belowcaptionskip}{0pt}
        \includegraphics[width=\textwidth]{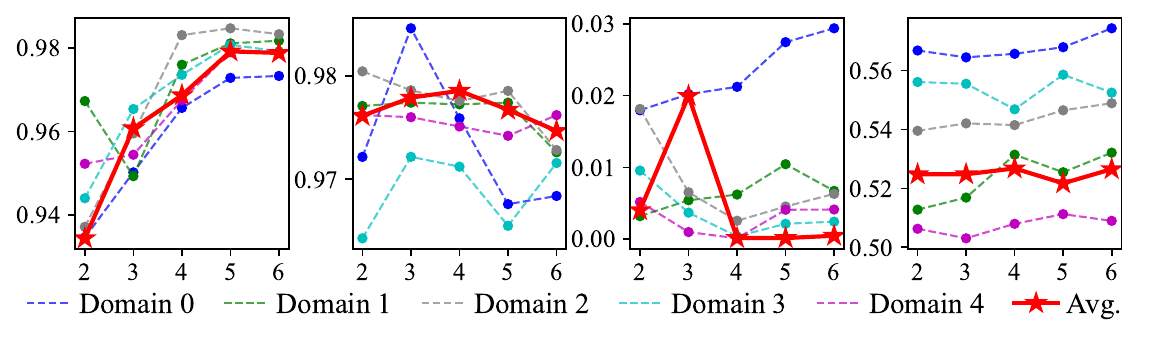}
            \caption{}
            \label{fig:nGMMs_fairface}
    \end{subfigure}
    \caption{Performance of \sysname{} on each domain and the average performance across different values of $K$ on (a) \texttt{RCMNIST}, (b) \texttt{NYPD} and (c) \texttt{FairFace} datasets. The x-axis is the values of $K$ ranging from 2 to 6.}
    \label{fig:nGMMs}
\end{figure}

\subsection{Sensitive Analysis}

\label{sec:trade-off}
\textbf{Accuracy-fairness Trade-off.} To assess the trade-off performance of \sysname{}, we obtained different group fairness and DG results of \sysname{} by controlling the value of $\lambda_2$ (larger $\lambda_2$ implies \sysname{} focuses more on algorithmic fairness). We compare the results with other fairness-aware methods, as shown in Figure \ref{fig:trade-off} for all three datasets. It can be seen that the curve of the results obtained by \sysname{} under different fairness levels is positioned in the upper-left corner among all methods. This indicates that \sysname{}, while ensuring the best fairness performance, also maintains comparable domain generalization performance, achieving the best accuracy-fairness trade-off. Moreover, we observe that \sysname{} achieves excellent fairness performance with comparable accuracy across all three datasets when $\lambda_2=0.5$. Therefore, we adopt this setting for all three datasets.

\textbf{Number of Prototypes.} To determine the number of prototypes $K$ in $g$, we conducted a sensitivity analysis of $K$. The experimental results on three datasets with fixed other parameters and varying values of $K$ from 2 to 6 are shown in Figure \ref{fig:nGMMs}. The number of prototypes we ultimately selected on the three datasets is 3, 3 and 4. Because at these values, \sysname{} had the highest average ranking across the four metrics as well as the best accuracy-fairness trade-off.

\section{Conclusion}
\label{sec:conclusion}
In this paper, we introduce a novel approach to fairness-aware learning that tackles the challenges of generalization from observed training domains to unseen testing domains. 
In our pursuit of learning a fairness-aware invariant predictor across domains, we assert the existence of an underlying transformation model that can transform instances from one domain to another. 
To ensure prediction with fairness between sensitive subgroups, we present a fair representation approach, wherein latent content factors encoded from the transformation model are reconstructed while minimizing sensitive information.
We present a practical and tractable algorithm. Exhaustive empirical studies showcase the algorithm's effectiveness through rigorous comparisons with state-of-the-art baselines.

\begin{acks}
This work is supported by the NSFC program (No. 62272338).
\end{acks}

\clearpage
\bibliographystyle{ACM-Reference-Format}
\balance
\bibliography{cikm24}

\appendix









\end{document}